\documentclass{ieeeaccess}
\usepackage{cite}
\usepackage{amsmath,amssymb,amsfonts}
\usepackage{algorithmic}
\usepackage{graphicx}
\usepackage{textcomp}
\usepackage{bm}
\usepackage{float}
\usepackage{color}
\usepackage{subcaption}
\usepackage{multicol}
\usepackage{siunitx}
\usepackage[hidelinks]{hyperref}
\graphicspath{ {./figures/} }
\def\BibTeX{{\rm B\kern-.05em{\sc i\kern-.025em b}\kern-.08em
    T\kern-.1667em\lower.7ex\hbox{E}\kern-.125emX}}
\newcommand\blfootnote[1]{%
  \begingroup
  \renewcommand\thefootnote{}\footnote{#1}%
  \addtocounter{footnote}{-1}%
  \endgroup
}

\begin{document}
\history{Date of publication 19 October 2022, date of current version 27 October 2022.}
\doi{10.1109/ACCESS.2022.3215984}

\title{Competitive Driving of Autonomous Vehicles}

\author{\uppercase{Gabriel Hartmann}\authorrefmark{1,2}, 
\uppercase{Zvi Shiller}\authorrefmark{1}, and \uppercase{Amos Azaria}\authorrefmark{2}}

\address[1]{Department of Mechanical Engineering and Mechatronics, 
Ariel University, Israel}
\address[2]{Department of Computer Science, 
Ariel University, Israel}
\address[]{gabrielh@ariel.ac.il, shiller@ariel.ac.il, amos.azaria@ariel.ac.il}
\tfootnote{This research was supported, in part, by the Ministry of Science \& Technology, Israel.}

\markboth
{Hartmann \headeretal: Competitive Driving of Autonomous Vehicles }
{Hartmann \headeretal: Competitive Driving of Autonomous Vehicles}
\corresp{Corresponding author: Gabriel Hartmann (e-mail: gabrielh@ariel.ac.il).}

\begin{abstract}
This paper addresses the issue of autonomous competitive yet safe driving in the context of the Indy Autonomous Challenge (IAC) simulation race. The IAC is the first multi-vehicle autonomous head-to-head competition, reaching speeds of 300 km/h along an oval track modeled after the Indianapolis Motor Speedway (IMS). 
We present a racing controller that attempts to maximize progress along the track while avoiding collisions with opponent vehicles and obeying the race rules. To this end, the racing controller first computes a race line offline.
During the race, it repeatedly computes a small set of dynamically feasible maneuver candidates, each tested for collision with the opponent vehicles.  
It then selects a collision-free maneuver that maximizes the progress along the track and obeys the race rules.  
Our controller was tested in a 6-vehicle simulation, managing to run competitively  with no collision over 30 laps.  
In addition, it managed to drive within a close range of the leading vehicle during most of the IAC final simulation race. 

\end{abstract}

\begin{keywords}
Autonomous vehicles, Collision avoidance, Motion planning, Multi-robot systems 
\end{keywords}

\titlepgskip=-15pt

\maketitle
\section{Introduction}v
\blfootnote{This work is licensed under a Creative Commons Attribution 4.0 License. For more information, see https://creativecommons.org/licenses/by/4.0/}

Autonomous racing has gained great interest in recent years \cite{racing_survey}, resulting in a number of racing competitions \cite{formula_student,roborace,INDY}. 
In competitive driving, the challenge is to maintain safety and minimize motion time while competing against other vehicles that attempt to achieve the same goal. 
We address the issue of competitive driving in the context of our participation in the recent Indy Autonomous Challenge (IAC) \cite{INDY}.
The IAC is an international competition intended to promote the development of algorithms for driving under challenging conditions. Its goal is to demonstrate the world's first multi-vehicle, high-speed, head-to-head autonomous racing.

\begin{figure}[h]%
    \centering
    \subfloat[\centering ]{{\includegraphics[height=2.05cm]{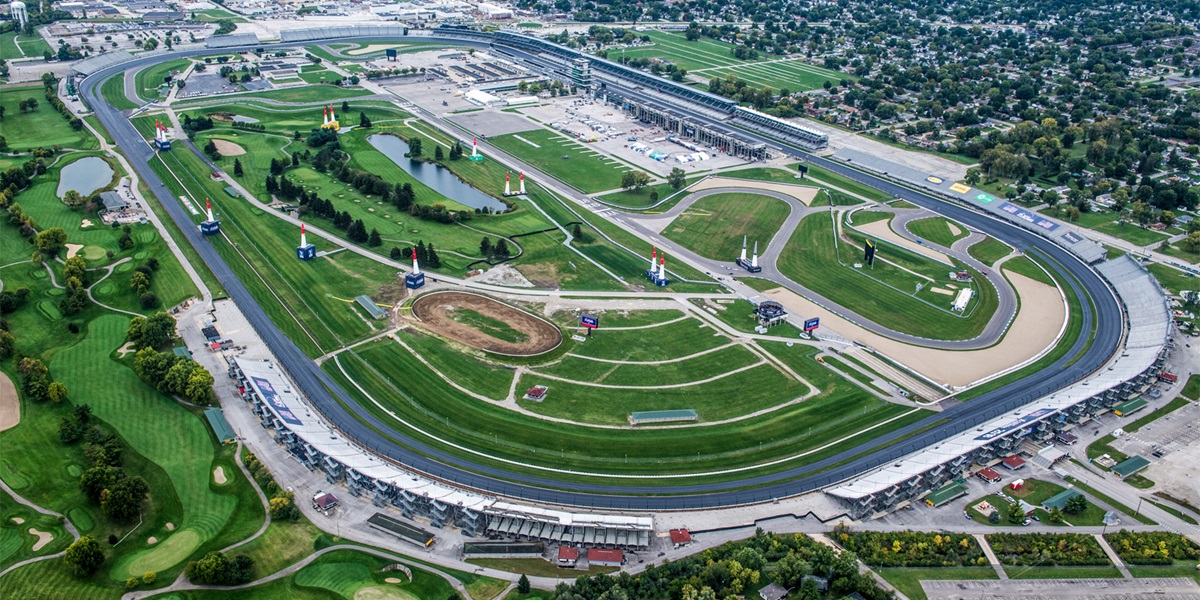} }}%
    \subfloat[\centering ]{{\includegraphics[height=2.05cm]{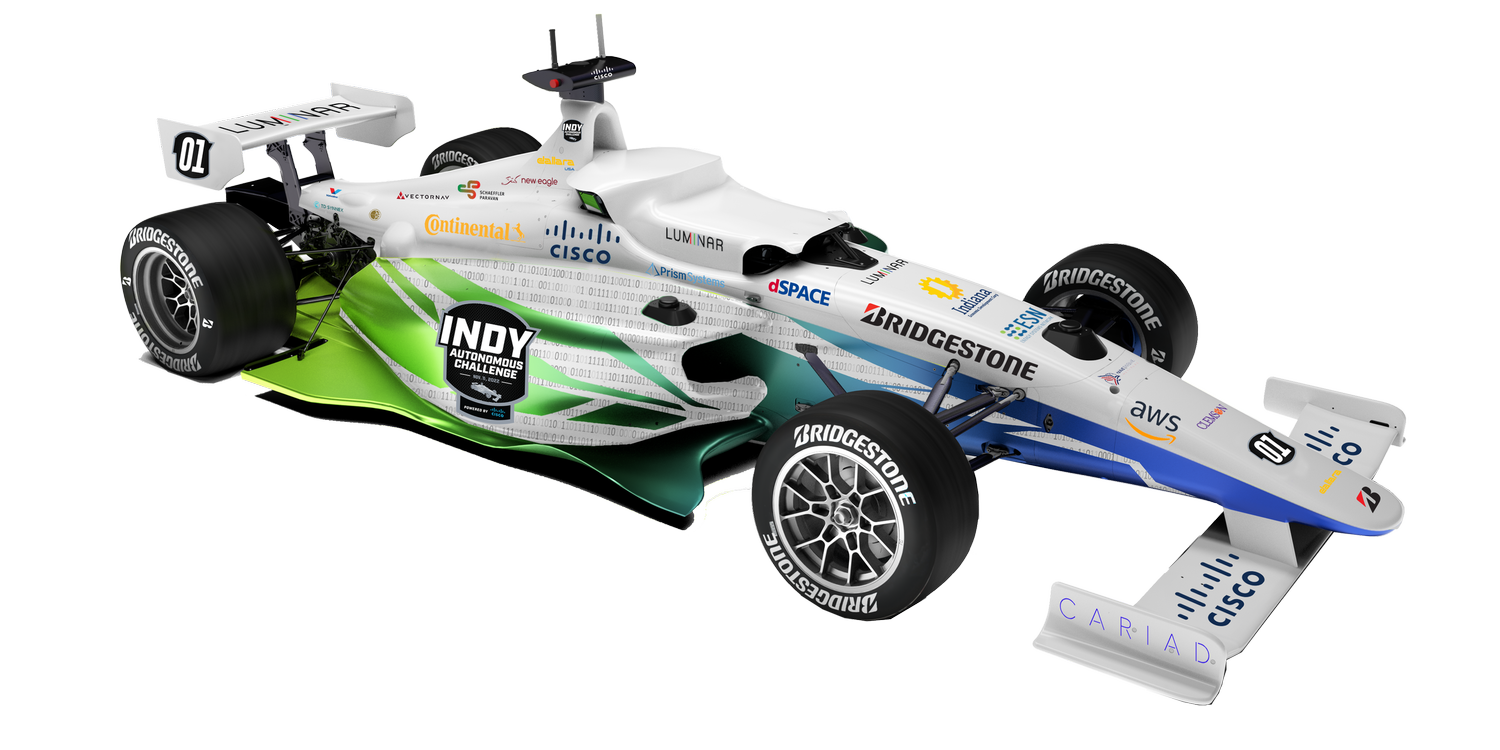} }}%
    \caption{(a) The Indianapolis Motor Speedway (b) the AV-21 autonomous race car \cite{INDY_racecar}}%
    \label{fig:ims_race_car}%
\end{figure}
The IAC, held on 2020-2021, was carried out in two stages: a simulation race and a real race on the Indianapolis Motor Speedway (IMS) with the Dallara AV-21 autonomous race car \cite{INDY_racecar}. Fig. \ref{fig:ims_race_car} depicts the IMS and the race car. 
Over $30$ teams from universities worldwide participated in this challenge.
A prerequisite for entering the competition was to demonstrate autonomous driving of a real vehicle.  Our team's entry submission is shown in  \cite{kia_demonstration}.

While the simulation race achieved its stated goals, the real race ended up with solo driving and controlled overtaking between two vehicles  \cite{INDY_ces}. 
Evidently, competitive driving with real vehicles is not yet ready and will hence rely for the time being on high-fidelity simulations. It is in the context of this reality that we present our work on competitive driving.    

\subsection{Challenges of autonomous racing}
Autonomous racing has unique challenges emanating from the unique properties of the race vehicle, its extreme speeds, and the competitive nature of the driving.    

\subsubsection{Extreme speeds}
Racing speeds coupled with limited frequencies of the sensor readings lead to state updates at large distance intervals compared to the vehicle size and the distance between neighboring vehicles.  Furthermore, driving near the vehicle's performance envelope in close proximity to neighboring vehicles leaves little room for correction and hence requires high-fidelity predictions of the behavior of the opponent vehicles.  

\subsubsection{Competitive driving} 
Competitive driving forces the competitors to race in close proximity to opponent vehicles.
As a result, the time difference between the leading teams is measured in fractions of a second. This, in turn, forces all competitors to drive on the performance envelope of the vehicle and the driver, leaving little room for safety.  
Although the goal is to win the race, in our opinion, especially for the first time that such a head-to-head race is taking place, safer behavior and larger safety margins should be preferred over pushing the vehicle's performance to its limits.  

\subsubsection{Aerodynamic forces}
\label{sec:Aerodynamics_forces}
The aerodynamics of a race car has two main effects: a down-force that increases the tire grip and consequently lets the car reach a lateral acceleration of over $2.5$ g, which in turn significantly increases the vehicle's maximal speeds along an oval track. The second effect is  slipstream, or drafting, which reduces the drag on the vehicle that follows behind at close range. Exploiting the slipstream is an important strategy in car racing since it allows vehicles with identical dynamics to overtake each other at high speeds. 

\subsubsection{Racing rules}
The racing rules \cite{INDY_rules} for the autonomous race were derived from the rules used in human-driven races \cite{Racerules}.   
Central to these rules is the principle that an overtaking vehicle is fully responsible for avoiding collision with a vehicle that is moving on its race line, without causing it to veer off its race line.  If a collision occurs in such a situation, the overtaking vehicle is held responsible for the collision and is removed from the race.  

\subsection{This paper}
The autonomous racing controller was developed based on the underlying principle that emphasizes safety over performance. To this end, our controller attempts to avoid collisions, including those that the race rules placed the responsibility to avoid on the opponent vehicle. While this is a conservative approach to competitive driving that attempts to drive safer than  what is allowed by the race rules, it is in our opinion the right approach to "responsible competitive driving."

The racing controller is based on a repeated search for the locally best maneuver that avoids collisions with opponent vehicles,  attempts to follow the globally optimal race line, and obeys the race rules.   

The best maneuver is selected from a tree of local maneuvers, generated to a set of goals across the track by applying time optimal control to a point mass model. The selected maneuver is then tracked using the pure-pursuit controller \cite{pure_pursuit}. 

Our controller was tested and analyzed in simulations with $6$ competing vehicles, all running the same algorithm. It was also tested in the IAC simulation race, where it competed in races with up to 7 vehicles.
We note that the controller is efficient for racing with even more vehicles since the algorithm's complexity is linear in the number of opponent vehicles.

Despite its simplicity, this controller demonstrated safe and competitive driving, while overtaking other vehicles over the $30$ lap run, without causing even a single  
collision between any of the $6$ competing vehicles. 
In the IAC simulation race, our vehicle managed to avoid collisions while staying within close range of the leading vehicle for the majority of the semi-finals and finals.  Our strategy of avoiding collisions at all cost  caused our vehicle to spin off the track while attempting to avoid collision with another vehicle that entered our safety bound. 
This placed our vehicle in the 6\textsuperscript{th} place in the simulation competition.  

The contributions of this paper are twofold:
\begin{itemize}
    \item Presenting a complete controller for a multi-vehicle race that drives competitively, avoids collision, and obeys the race rules.
    \item Presenting multi-vehicle ($6$ vehicles) race results and providing metrics to quantify the performance of a multi-vehicle race.
\end{itemize}

\section{Related Work}
Most previous work in the field of autonomous racing have focused on solo-racing \cite{Hierarchical_Motion_Planning,kabzan2020amz,RRT_halfcar} or racing against a single opponent vehicle \cite{AutoRally_differential_games,liautonomous}. Very few studies have addressed multi-vehicle racing with more than two opponent vehicles \cite{Game_Theoretic_3_vehicles,wurman2022outracing}.
This paper describes our multi-vehicle racing controller and demonstrates it (in simulation) in a 6-vehicle and 7-vehicle races.

Autonomous racing has been tested in competitions, such as the Formula student challenge \cite{formula_student} or Roborace \cite{roborace}, which focus mainly on solo racing. The recent IAC real race demonstrated controlled overtaking between two vehicles   \cite{INDY_ces}. A multi-vehicle race has been so far demonstrated only in the IAC simulation race \cite{INDY}.
For a comprehensive survey on autonomous racing, see \cite{racing_survey}.

We now survey some of the existing work on autonomous racing regarding each of our racing algorithm modules, namely global race line planning, opponents' trajectories prediction, local planning, and trajectory tracking.

\paragraph{Global race line planning}
A global race line is the time optimal path (the projection of the time optimal trajectory on the track) for a one lap solo race. Computed offline, it serves as a reference input to   
a trajectory-following controller \cite{Hierarchical_Motion_Planning,Adaptive-pure-pursuit}. It is also used 
in multi-vehicle racing as a reference for the online planner, as is demonstrated later in this paper. 

The race line optimization for autonomous racing usually minimizes lap time, which can be computed  by solving a nonlinear optimization problem that considers vehicle dynamics and track constraints \cite{Hierarchical_Motion_Planning,lap_optimization}. An alternative approach is to construct a race line by minimizing curvature along the path, which impacts the vehicle's lateral acceleration \cite{Minimum_curvature}. Since vehicle speeds vary little along the oval track, optimizing path curvature offers a good kinematic approximation for the  time optimal dynamic optimization.  It is computationally efficient and it produces race lines that differ only slightly from the time optimal line. 
We used open source software \cite{TUMFTM} for the offline race line optimization.

\paragraph{Trajectory prediction}
An early approach to trajectory prediction was based on Kalman filter to generate short time horizon predictions while assuming a constant velocity and acceleration along each segment \cite{ammoun2009real}. A more recent approach trains recurrent neural networks to predict the multi-modal distribution of future trajectories \cite{deo2018multi}.
While the learning-based approach, using neural networks for prediction, is promising, it requires a large amount of data, and it does not guarantee the quality of the learned solution \cite{Structured_Prediction}. 

For computational efficiency, we devised a prediction module that accounts for the opponent's current state and track geometry. We used a conservative assumption that while moving across the  track, the opponent vehicle will cross the entire track, until reaching either the inner or outer boundary, and will follow that track boundary while maximizing its speed during the set time horizon.  

\paragraph{Local planning}
Model Predictive Control (MPC) was used to optimize a local trajectory for solo driving \cite{kabzan2020amz,hessian_MPC}, and for avoiding static  \cite{liniger2015optimization} and  dynamic obstacles \cite{Tube_MPC}. To avoid obstacles, Wischnewski et al. \cite{Tube_MPC} and Liniger et al. \cite{liniger2015optimization} first plan a free driving zone using graph search, then generate an optimal trajectory within the selected zone. 
In \cite{liautonomous}, MPC was used directly to overtake a single opponent vehicle in a racing scenario.

Random sampling was used to search for a feasible, collision-free solution using Rapid Random Trees (RRT) \cite{RRT} and  RRT* \cite{RRT_halfcar} and CLRRT\textsuperscript{\#} \cite{CLRRT} to compute an asymptotically optimal solution.  
It is important to note that random sampling-based methods do not guarantee a solution in a finite computation time. Furthermore, these methods typically require excessive computation time that is detrimental to real-time execution at high-frequency. 

A less computationally demanding solution is to select an optimal trajectory from a small set of trajectories, generated, for example, by simulating various constant steering angles and velocities \cite{liniger2015optimization}, or by generating cubic spirals to a set of target points \cite{TuneCar}.

The above mentioned methods are either unsuitable for high-speed racing or have not been demonstrated for more than three vehicles. 
Our online planner selects a trajectory from a small set of maneuvers generated by a point mass model to a number of target points, while minimizing motion time, avoiding collisions with the opponent vehicles, and attempting to reach the optimal race line, as described later in Section \ref{sec:trajectory_planning}.

\paragraph{Trajectory tracking}
Trajectory tracking can be done using feedback control by minimizing position error,  \cite{Adaptive-pure-pursuit,ggdiagram_control}, 
Model Predictive Control (MPC) by generating a sequence of open loop commands that minimize tracking error subject to dynamic constraints \cite{Tube_MPC,Hierarchical_Motion_Planning}, and learning-based control that iteratively minimizes lap time and tracking error \cite{hartmann2019deep, rosolia2017autonomous}.

We opted for the pure-pursuit algorithm \cite{pure_pursuit}, together with low-level linear and angular velocity controllers.  It provides satisfactory performance when fine-tuned for driving along the smooth trajectories generated by the local planner. 
It is easy to tune, owing to the low number of parameters used, compared to other methods.

\paragraph{Reinforcement learning approach} %

Recently, Wurman et al. \cite{wurman2022outracing} beat human drivers in a racing video game, using reinforcement learning. Unlike the hierarchical approach of the methods described earlier, the reinforcement learning agent maps a low-dimensional state directly to control commands.
Although learning-based approaches are promising, they usually require a high amount of training data, making it challenging to generalize them to training safely in the real world. Such long training also made it impractical for the IAC simulation race because of the slower-than-real-time performance of the simulator. 
Another challenge is to validate the safety of the learned controller, which is nontransparent (i.e., black box), unlike physics-based controllers.

\paragraph{Comparison}
Most of the current research focuses on single or two-vehicle racing, and there is currently no demonstration of a real race with more than two vehicles. We note that it is impossible to directly compare different methods implemented in different settings in the racing domain. This is because the performance differences between racing competitors are very small, as shown later in Section \ref{sec:Experiments}. In addition, in a racing environment, all methods are extensively fine-tuned to a specific setting, and thus, one method cannot be applied, as is, to a different setting. These challenges emphasize the importance of racing competitions such as the Indy Autonomous Challenge that provides an opportunity to compare the performance of various methods on a common ground.

\section{Software Architecture}
\begin{figure}[h]
    \centering
    \includegraphics[width=\linewidth]{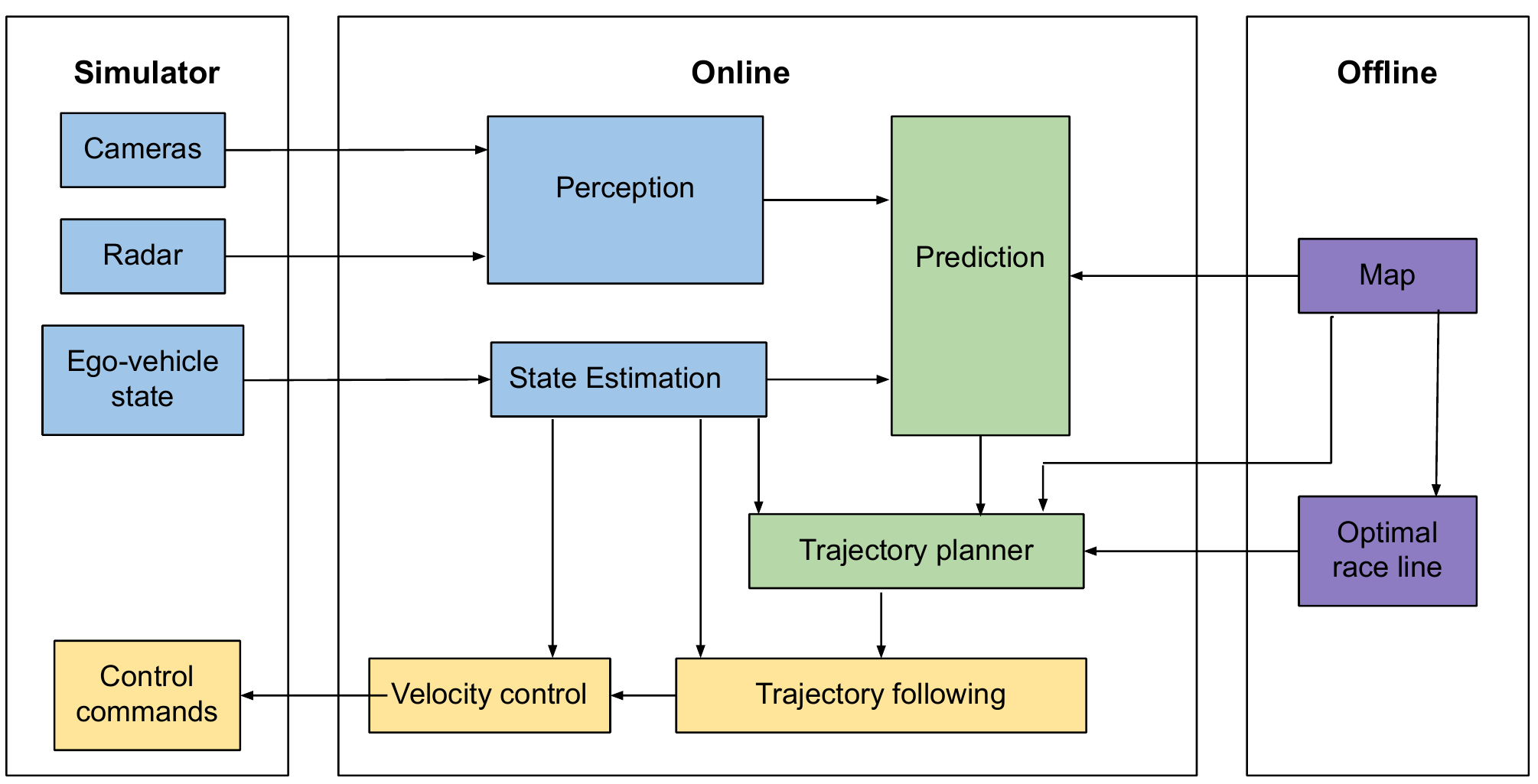}
    \caption{The software architecture.}
    \label{fig:architecture}
\end{figure}

The software architecture of the racing controller is shown schematically in Fig. \ref{fig:architecture}. An optimal race line is computed offline for a given track, which serves the controller throughout the race. 
The data from the cameras and radars provide the position and velocity of the surrounding opponent vehicles, and additional simulated sensors provide the ego-vehicle state, e.g., position and velocity. Together with the map and the optimal race line, these data are used by the prediction module to predict the opponent vehicles' future trajectories repeatedly. The trajectory planner uses the same information to plan an optimal local maneuver for the ego-vehicle. 
This maneuver serves as an input to the trajectory-following controller, which computes the ego-vehicle's desired linear and angular velocities. The linear and angular velocities are controlled by the velocity controller, which outputs the steering, throttle, and brake commands.

\section{Computing the optimal race line}
The optimal race line is a time-optimal trajectory when no other vehicles are present on the track. It is computed offline, based on the track geometry and vehicle dynamics.

The IMS  is an oval track, $4,023$ m ($2.5$ miles) long, as depicted in Fig. \ref{fig:map}.

We computed the optimal race line using an open-source trajectory optimization software  \cite{TUMFTM}.  
The optimal race line typically minimizes curvature by entering the corner on the outside boundary of the track, passing through the apex on the inner boundary to the exit point on the outside boundary, as shown in Fig. \ref{fig:corner}.

\newcommand{\xmap}{4.3cm}
\begin{figure}[h]%
    \centering
    \subfloat[\centering ]{{\includegraphics[height=\xmap]{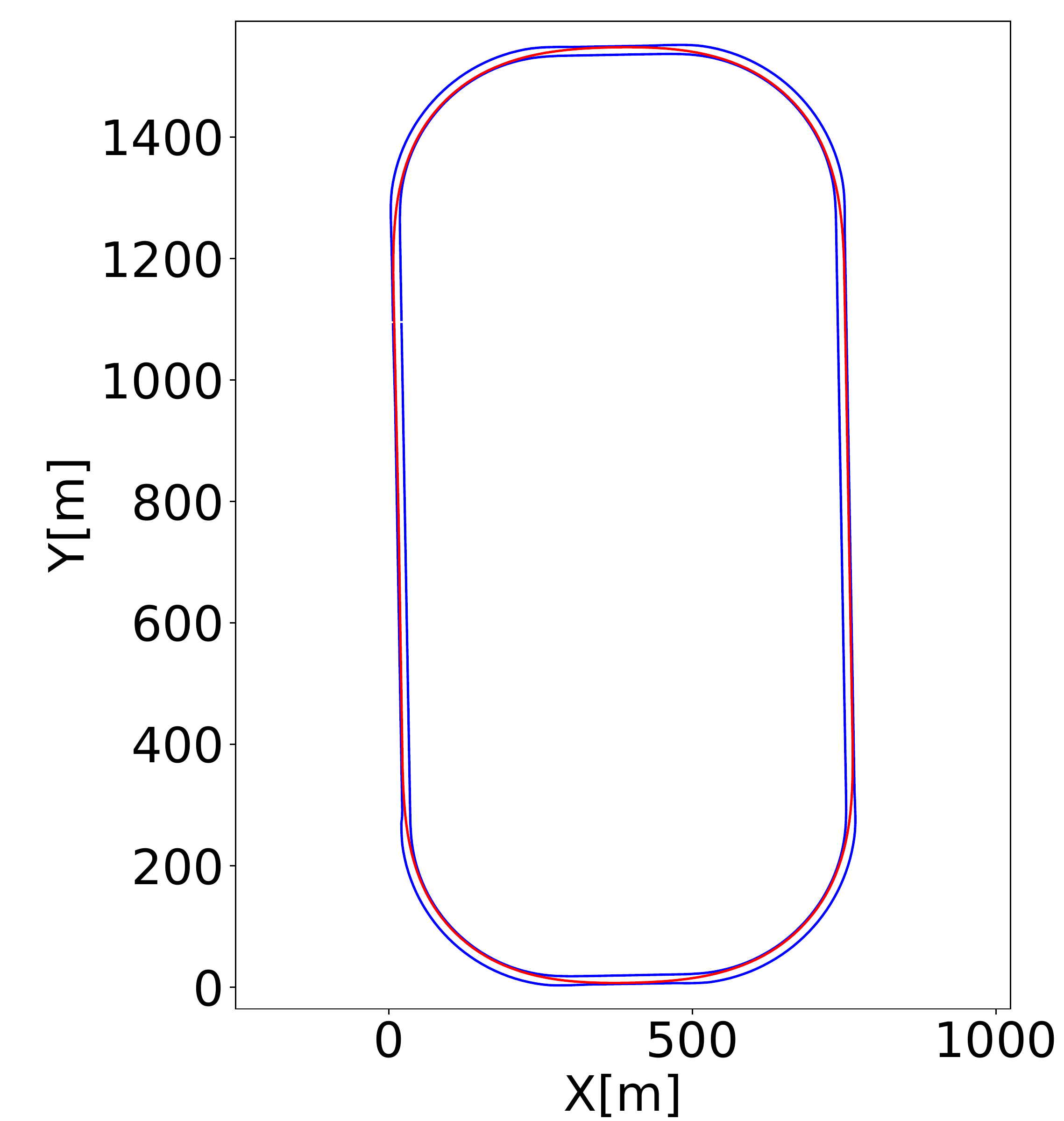} }\label{fig:map} }%
    \subfloat[\centering]{{\includegraphics[height=\xmap]{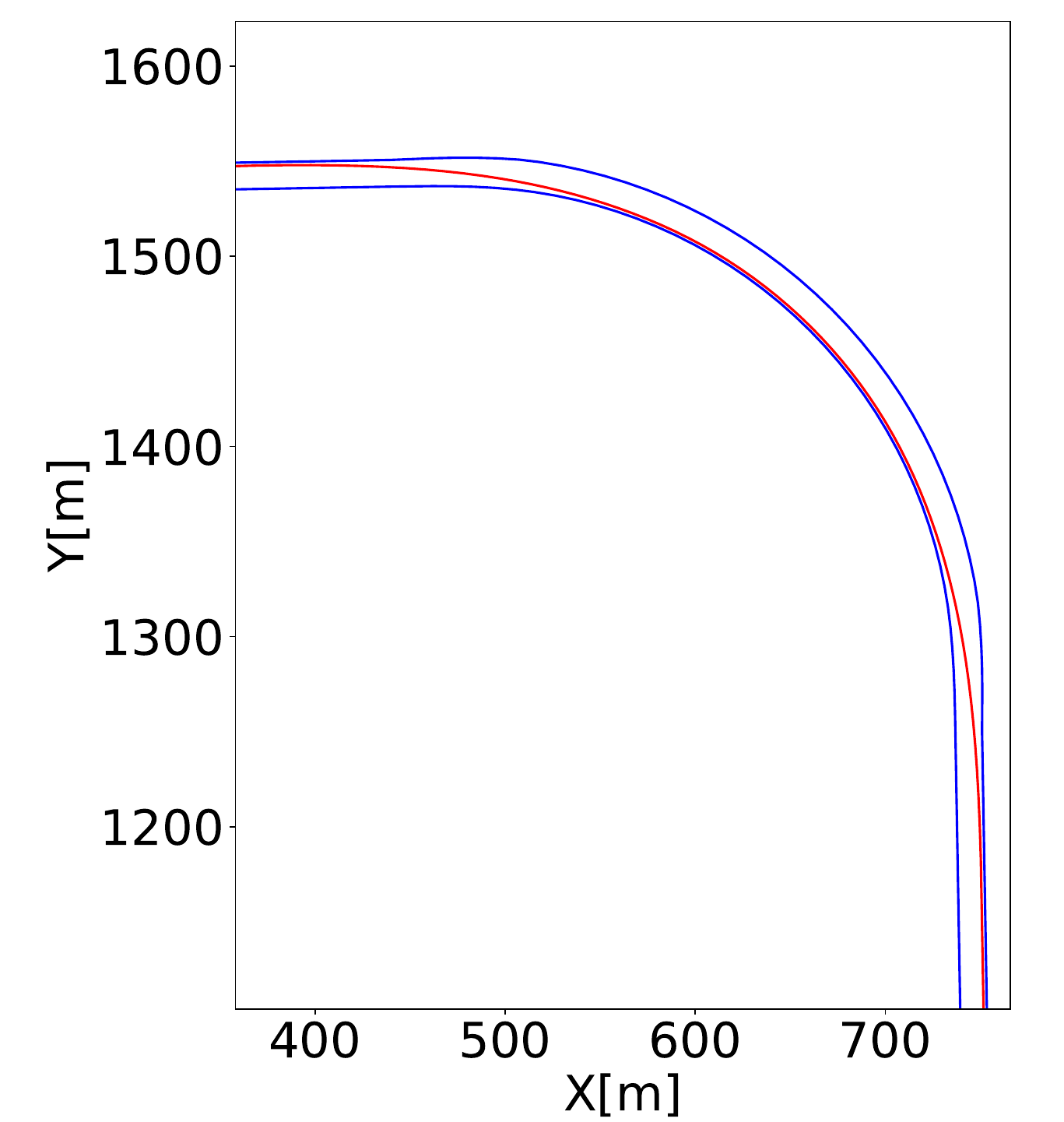} }\label{fig:corner}}%
    \caption{(a) The IMS track. (b) The optimal race line around a corner of the track.}%
    \label{fig:offline_trajectory}%
\end{figure}

\section{Online Trajectory Planning}
\label{sec:trajectory_planning}
The online trajectory planner computes a collision-free trajectory at $25$ Hz. This rate was dictated by the frequency of the simulator sensors updates.  
This in turn resulted in a travel distance of $3$ m between trajectory updates.

The trajectory computed by the online planner is used as a reference for the trajectory-following controller (see Section \ref{sec:control}). 
The planner first generates a tree of dynamically feasible maneuver candidates to $8$ points that span the track width.  
The best candidate that maximizes progress along the path and avoids collision with the opponent vehicles is then selected. 

The online trajectory planner uses a horizon of $200$ m, which the stopping distance at $80$ m/s.  This is also compatible with the sensor range of $200$ m. %
The limited planning horizon decreases the optimality of the local trajectory. However, we mitigate this by attempting to merge with the optimal race line, which is in itself  globally optimal.
Despite the short time horizon, we demonstrate that the vehicle is able to drive safely and 
competitively, as described later in Sec. \ref{sec:Experiments}.

\subsection{Coordinate system}
\begin{figure}[H]
    \centering
    \includegraphics[width=0.7\linewidth]{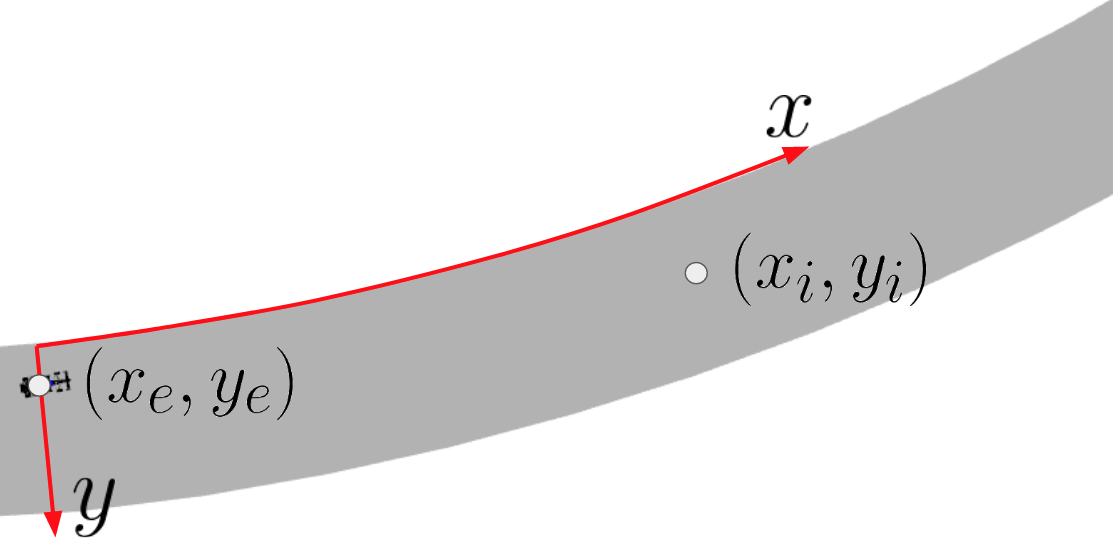}
    \caption{Road-aligned coordinate system.}
    \label{fig:coordinates}
\end{figure}

The coordinate system  used in planning the trajectory is attached to the ego-vehicle  with the $x$-axis tangent to the left track boundary 
and the $y$-axis normal to track. The position of the ego-vehicle is denoted  $(x_e,y_e)$.   The 
ego-vehicle is always located at $x_e = 0$, and $y_e$ represents the ego-vehicle's normal offset from the left boundary (see Fig. \ref{fig:coordinates}).
In our coordinate system, for every point $(x_i,y_i)$, $|x_i|$ represents the distance from the ego-vehicle along the track (a negative $x_i$ represents a point behind the ego-vehicle), and $y_i$ represents the offset from the track boundary regardless of the track shape and the ego-vehicle's location.

\subsection{Point-to-point trajectory }%
\label{maneuver_planning}
We now present the computation of a  trajectory between two end points, for a point mass model.   
This  is repeatedly used to predict the trajectories of  opponent vehicles   and for planning maneuver candidates for the ego-vehicle. 

Let $p_s = \{x_0,y_0,\dot{x}_0,\dot{y}_0\}$ be the starting point, and  $p_g = \{x_g,y_g,\dot{x}_g,\dot{y}_g\}$ be the goal point.
We wish to plan a trajectory $C(t) = \{x(t),y(t),\dot{x}(t),\dot{y}(t)\}, \ t\in[0,T]$ so that  $C(0)=p_s$ and $C(T)=p_g$. We assume a constant longitudinal speed (along the $x$ axis) so that $\dot{x}_s=\dot{x}_g$.  
The travel time $T$ to the goal is therefore   
$T = (x_g - x_0)/\dot{x}_0$.   
Since  the longitudinal distance to the goal  is by choice greater than the track width,  
$C$ is thus generated by 
using bang-bang control in the lateral direction, while applying the minimum lateral force with a single switch so that the lateral and the longitudinal end points would be reached both at time $T$
\cite{shiller1998emergency}. 

The lateral force $F_y$  that generates the bang-bang trajectory 
is:     
\begin{equation} 
F_y =  \frac{-\sqrt {2A}-T(\dot{y}_g + \dot{y}_0)+2(y_g - y_0)}{T^2m}   
\end{equation}
where
\begin{equation}
A = (T^2  (\dot{y}_0^2 + \dot{y}_g^2) -2T(y_g - y_0)(\dot{y}_g + \dot{y}_0)+2{(y_g - y_0)}^2), 
\end{equation}
and the switching time $T_s$ is
\begin{equation}
T_{s} = \frac{ m(\dot{y}_g - \dot{y}_0) +F_yT}{2F_y(t)}.
\end{equation}
$C(t)$ is thus computed for $t \in[0,T]$ by:
\begin{align} 
x(t) &= x_0 + \dot{x}_0 t \\
y(t) &= \left\{ \begin{matrix}
   y_0 + \dot{y}_0t + \frac{1}{2}\frac{F_y}{m}t^2 \\ \\
   y(T_s) + \dot{y}(T_s)(t-T_s) - \frac{1}{2}\frac{F_y}{m}(t-Ts)^2 \\
\end{matrix}\begin{matrix}
   \ \ \  t \leq T_{s}  \\ \\
   \ \ \  t > T_{s}  \\
\end{matrix} \right.  \\
\dot{x}(t) &= \dot{x}_0 \\
\dot{y}(t) &= \left\{ \begin{matrix}
   \dot{y}_0 + \frac{F_y}{m}t \\ \\
   \dot{y}_0 + \frac{F_y}{m}(2T_{s} - t) \\
\end{matrix}\begin{matrix}
   \ \ \  t \leq T_{s}  \\ \\
   \ \ \  t > T_{s}  \\
\end{matrix} \right.  
\end{align}

Figure \ref{fig:point_mass_trajectory} illustrates an example of a trajectory between two given end states.

\newcommand{\x}{2.4cm}
\begin{figure}[h]%
    \centering
       \subfloat[\centering ]{{\includegraphics[height=2.0cm]{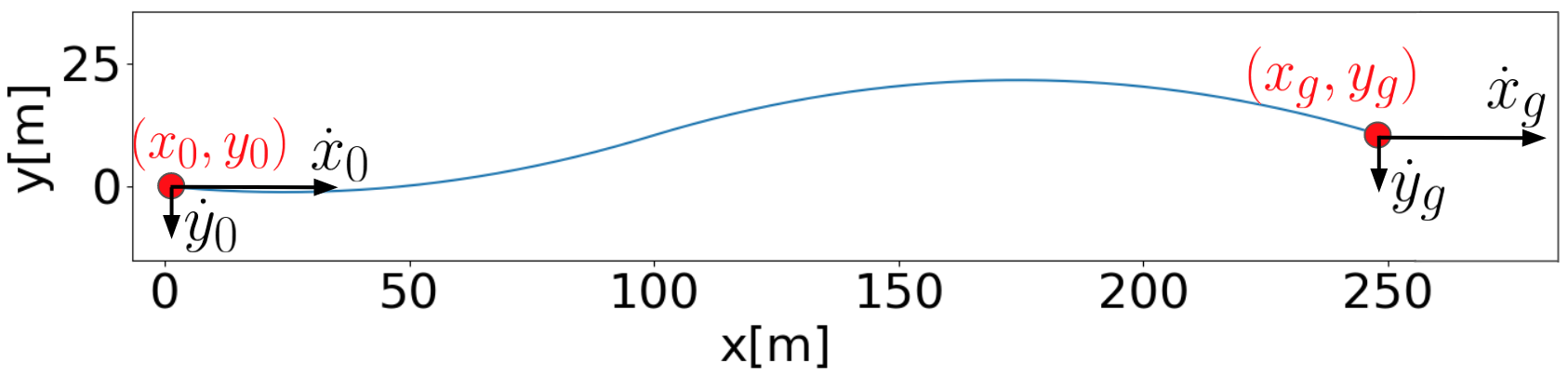} }}%
    \qquad
    
    \subfloat[\centering ]{{\includegraphics[height=2.65cm]{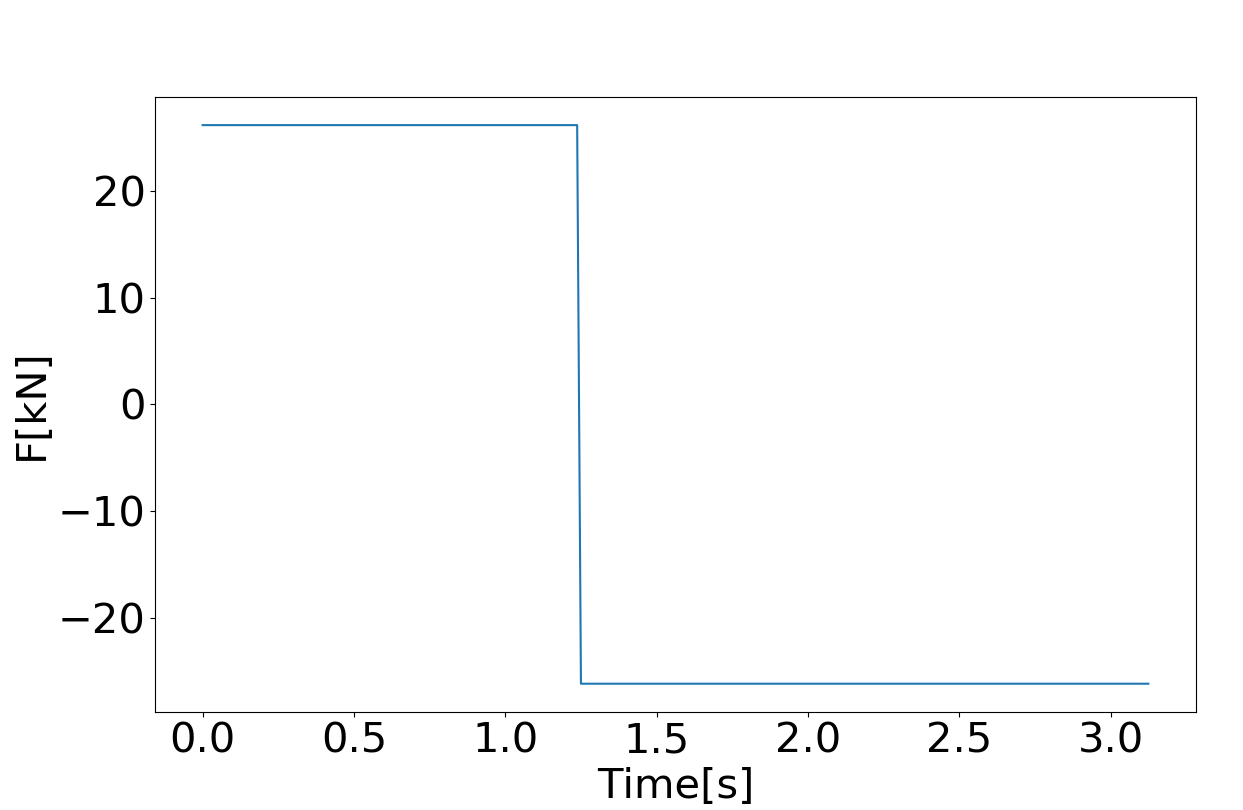} }}%
    \subfloat[\centering ]{{\includegraphics[height=\x]{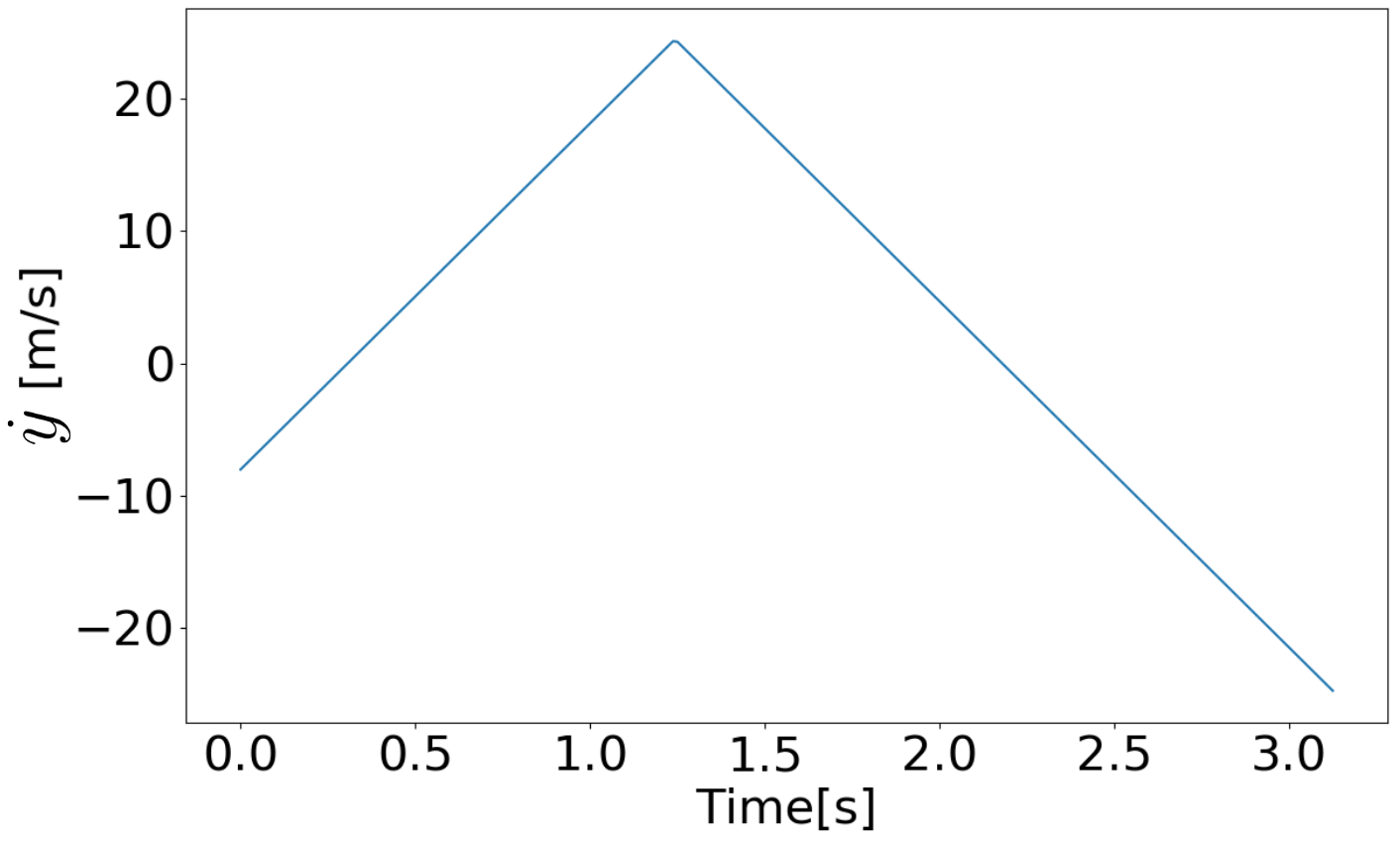} }}%
    \qquad
    \caption{A trajectory between given start and goal points. (a) Force $F_y$ is applied on mass $m$ to create a continuous trajectory between the points. (b) The lateral force $F_y$ profile and (c) the lateral velocity $\dot{y}(t)$.}%
    \label{fig:point_mass_trajectory}%
\end{figure}

\subsection{Trajectory prediction of opponent vehicles }
\label{sec:prediction}
An important part of motion planning in a dynamic environment, especially in racing, is predicting the future positions and velocities of all other vehicles surrounding the ego-vehicle. This is performed by the prediction module, which receives the track boundaries and an opponent vehicle state $s=\{x_0,y_0,\dot{x}_0,\dot{y}_0,\omega_0\}$ as input, where  $x_0,y_0$ is the opponent vehicle's position, $\dot{x}_0,\dot{y}_0$ is its velocity, and  $\omega_0$ is its angular velocity. The module predicts the future trajectory $J(t) = \{x(t),y(t), \dot{x}(t),\dot{y}(t) \}$ up to a predefined time horizon $T_{\textrm{max}}$, which was set to $3$ seconds to match the planning horizon. 

The prediction module predicts each opponent's trajectory based on its current state under the following two assumptions: the opponent vehicle intends to stay within the track boundaries (as described earlier) and attempts to maximize its velocity (as described in Section \ref{sec:re-planning}).
Namely, for a given opponent vehicle, we first predict a future trajectory, $\hat{J}$, that keeps a constant curvature $\kappa$, which we approximate by:
$\kappa = \frac{\omega_o } {||\dot{x}_o+\dot{y}_o||}$.

It is assumed that the opponent vehicle will stay within the track boundaries. 
to this end, our prediction module accounts for the track boundaries as follows:
Let $(\hat{x},\hat{y})$ be the first position on $\hat{J}$, at which the opponent vehicle approaches one of the track boundaries at a distance  $d_{\textrm{min}}$. 
We define three points, $p_0$, $p_1$ and $p_2$,  each point consisting of $p_i=\{x_{p_i},y_{p_i},\dot{x}_{p_i}, \dot{y}_{p_i}\}$. The prediction module connects these points by a point mass maneuver as explained in Section \ref{maneuver_planning}. 

The first point, $p_0$ is derived from the opponent vehicle's current state $s$, such that $x_{p_0} = x_o$ ,$y_{p_0} = y_o$, $\dot{x}_{p_0} = \dot{x}_o$, and $\dot{y}_{p_0} = \dot{y}_o$. The second point, $p_1$, is based on $(\hat{x},\hat{y})$, but we assume that the opponent vehicle will not increase its curvature; therefore, we assume that $\hat{y}$ will be reached later on, by a predefined factor, $k$. That is, $x_{p_1} = k(\hat{x}-x_o), y_{p_1} = \hat{y}, \dot{x}_{p_1} = \dot{x}_o, \dot{y}_{p_1} = 0\}$. %

\begin{figure}[h]
    \centering
    \begin{subfigure}{\linewidth}
        \centering
        \includegraphics[width=0.8\linewidth]{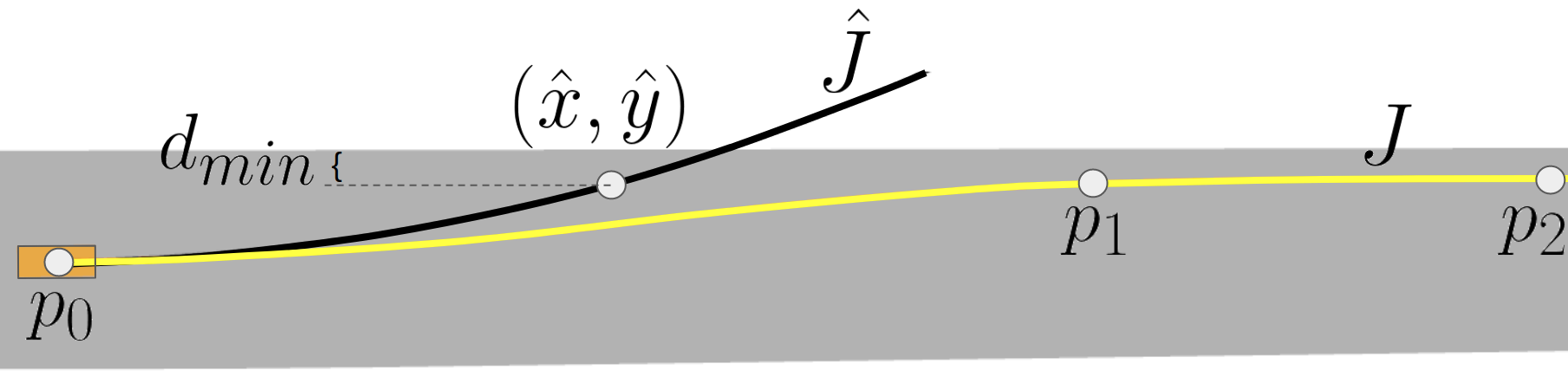}
        \caption{}
        \label{fig:prediction_change_lane}
    \end{subfigure} 
     \begin{subfigure}{\linewidth}
        \centering
        \includegraphics[width=0.8\linewidth]{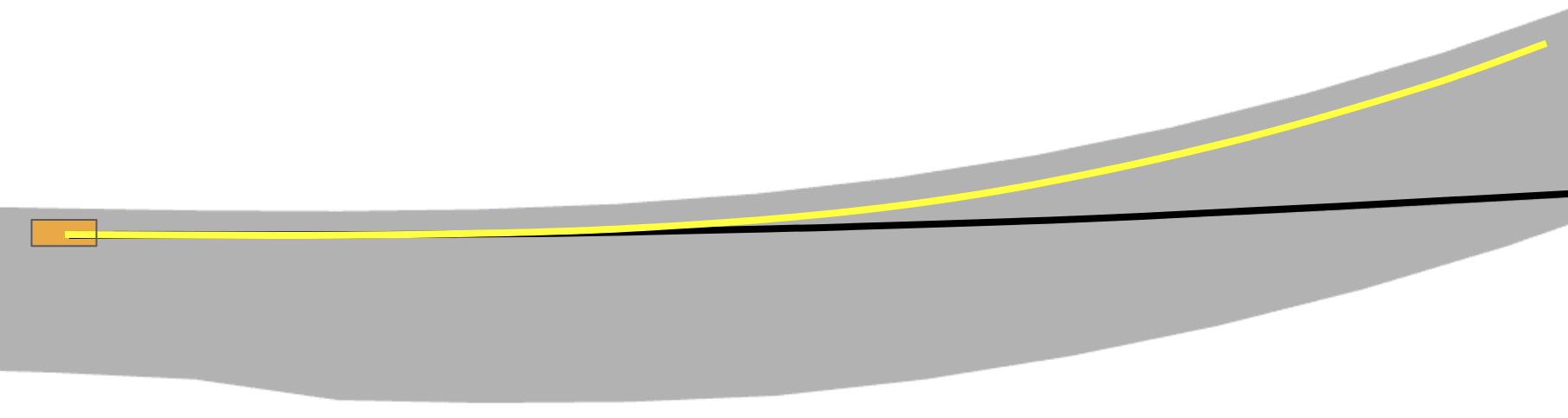}
        \caption{}
        \label{fig:prediction_follow_curve}
    \end{subfigure} 
\caption{Examples of predicted trajectories of an opponent vehicle (marked as an orange bounding box). Black indicates constant curvature trajectory $\hat{J}$, and yellow indicates the predicted trajectory $J$, which also considers the track boundaries. In (a) $\hat{J}$ exceeds track boundaries; therefore, a lateral-shift trajectory is predicted by $J$; in (b), although the vehicle drives on a straight line, the predicted trajectory  $J$ follows the track geometry.}
\label{fig:prediction}
\end{figure}

The third point, $p_2$ retains a path parallel to the boundary, up to $T_{\textrm{max}}$, i.e., $x_{p_2} = \dot{x}_o T_{\textrm{max}}$, $y_{p_2} = \hat{y}$, $\dot{x}_{p_2} = \dot{x}_o$, and $\dot{y}_{p_2} = 0\}$. 
Examples of predicted trajectories are shown in Fig. \ref{fig:prediction}.

\subsection{Creating maneuver candidates}
The online trajectory planner plans a set of dynamically feasible maneuver candidates and selects one according to multiple criteria, as follows:

\subsubsection{Lateral-shift maneuver}
\label{sec:lane_changing_maneuver}
\begin{figure}[h]
    \centering
    \begin{subfigure}{0.5\linewidth}
        \centering
        \includegraphics[width=\linewidth]{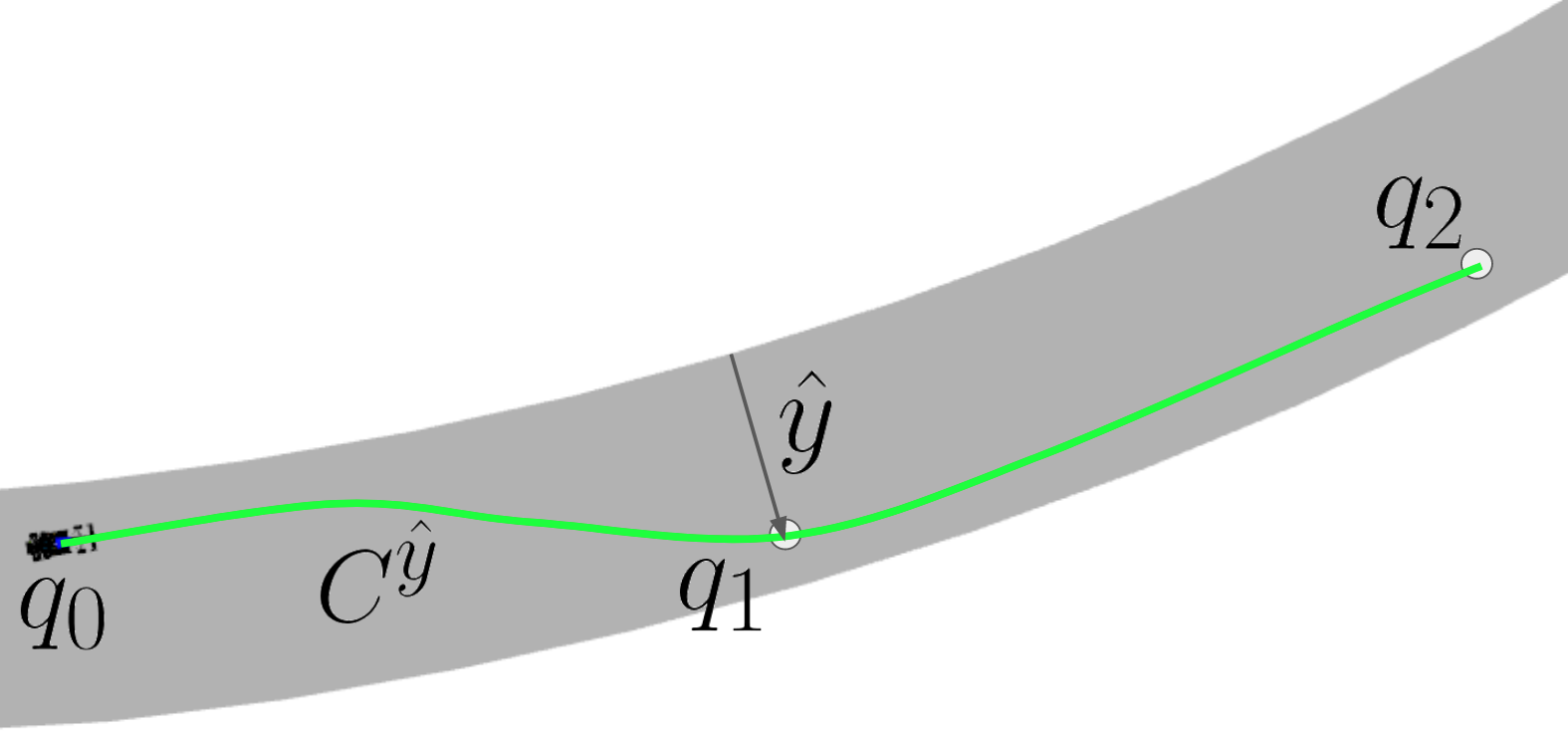}
        \caption{}
        \label{fig:change_lane}
    \end{subfigure}%
    \begin{subfigure}{0.5\linewidth}
        \centering
        \includegraphics[width=\linewidth]{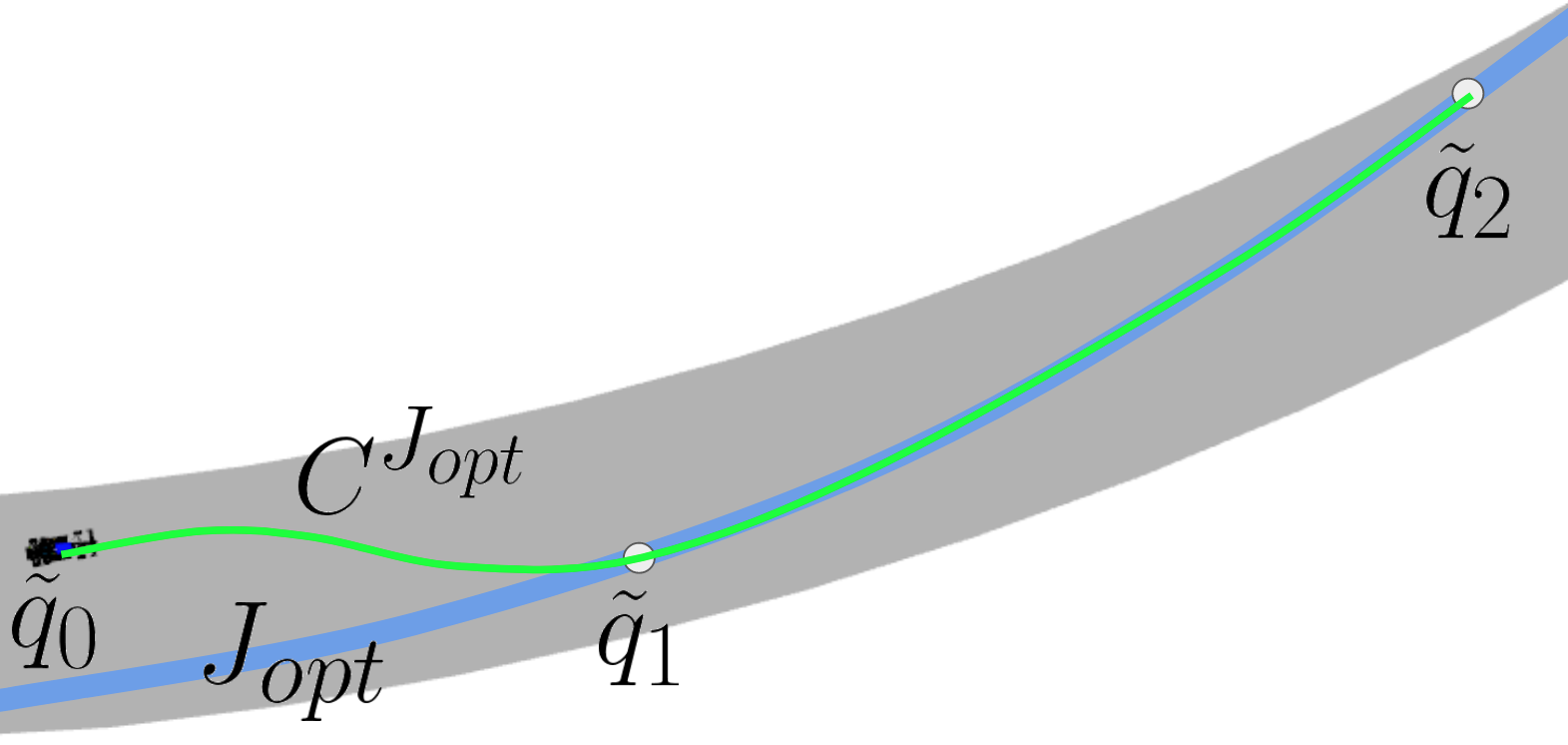}
        \caption{}
        \label{fig:to_optimal}
    \end{subfigure} 
\caption{Illustration of the maneuver candidates planning: (a) lateral-shift maneuver and (b) a maneuver to the optimal race line. }
\label{fig:maneuver_planning}
\end{figure}

Given the ego-vehicle's state $s_e=\{x_e,y_e,\dot{x}_e,\dot{y}_e\}$ and a lateral-shift target $\hat{y}$, the planner module creates a lateral-shift maneuver, $C^{\hat{y}}$, by connecting three points, 
$q_i=\{x_{q_i},y_{q_i},\dot{x}_{q_i}, \dot{y}_{q_i}\}$, $i\in \{0,1,2\}$, 
consisting of the $q_0$, the ego vehicle's current state, $q_1$ being the target point of the lateral-shift, and $q_2$ being a point down the track that retains the lateral shift of $q_1$.  
More formally, 
$q_0 = s_e$,
$q_1 = \{(\hat{y} - y_e) b + c,\hat{y},\dot{x}_e,0\}$, 
$q_2 =\{x_\textrm{max},\hat{y},\dot{x}_e,0\}$,
where $\hat{y}$ is the lateral-shift target,  $b,c$ are constants and $x_\textrm{max}$ is the planning horizon.
Figure \ref{fig:change_lane} illustrates a lateral-shift maneuver. 

We define $N$ equally spaced lateral-shifting targets representing potential $\hat{y}$ values. The first and last targets are far enough from the boundaries to allow a vehicle to reach them safely.
Our planner generates $N$ lateral-shift maneuvers, one for each defined target. Let $\boldsymbol{M'}$ be the set of all lateral-shift maneuvers:
\begin{equation}
   \boldsymbol{M'} = \bigcup\limits_{i=0}^{N-1} C^{{\hat{y}}_i}, \ \hat{y}_i = d_{\textrm{min}}+\frac{w - 2d_{\textrm{min}}}{N-1}i
\end{equation}
where $d_{\textrm{min}}$ is the minimal distance to keep away from track boundaries.
The width of the track, $w$, in most segments along the oval is $14$ m; we set the number of lateral-shift maneuvers $N = 7$ to achieve a distance between targets that is close to the vehicle's width ($2$ m ).

\subsubsection{Maneuver to the optimal race line}
In addition to these lateral-shift maneuvers, we plan a maneuver ${C}^{J_{\textrm{opt}}}$, which smoothly merges with the optimal race line $J_{\textrm{opt}}$.
We define the following three points $\tilde{q}_i=\{x_{\tilde{q}_i},y_{\tilde{q}_i},\dot{x}_{\tilde{q}_i}, \dot{y}_{\tilde{q}_i}\}$, where $i\in \{0,1,2\}$.

The first point $\tilde{q}_0 = s_e$; the second, $\tilde{q}_1\in J_{\textrm{opt}}$ such that $(y_{\tilde{q}_1} - y_e) \tilde{b} + \tilde{c} = x_{\tilde{q}_1}$, where $\tilde{b}$ and $\tilde{c}$  are predefined constants. Finally, $\tilde{q}_2\in J_{\textrm{opt}}$ such that $x_{\tilde{q}_2} = x_{\textrm{max}}$. See Fig.  \ref{fig:to_optimal}).

\begin{figure}[h]
    \centering
    \includegraphics[width=\linewidth]{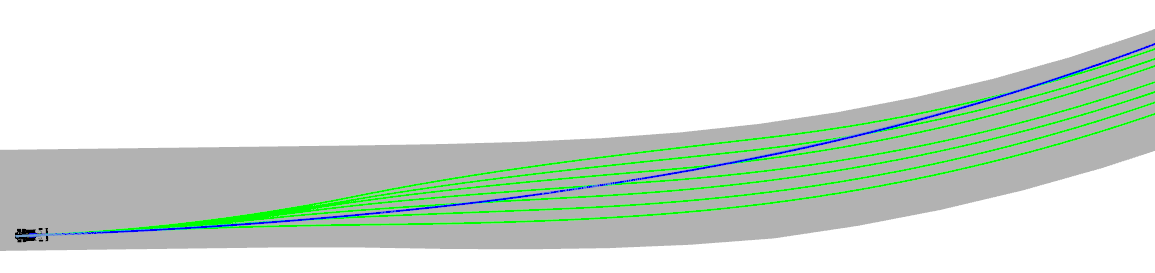}
    \caption{Example of $7$ lateral-shift maneuvers (green) and a maneuver that merges with the optimal race line (blue). }
    \label{fig:optional_trajectories}
\end{figure}

The full set of the maneuver candidates is $\boldsymbol{M} = \boldsymbol{M'} \cup C^{J_{\textrm{opt}}}$, as shown in Fig. \ref{fig:optional_trajectories}.

\subsection{Velocity re-planning}
\label{sec:re-planning}
Although the maneuver candidates $\boldsymbol{M}$ and the predicted trajectories of the opponent vehicle include velocities in addition to positions, these velocities were only used to define the direction of the paths and to estimate the lateral forces on the vehicle.
Therefore, the velocity profiles are re-planned to represent the future motion more accurately by assuming that the vehicles accelerate along the trajectories until they reach the maximal velocity.
This is possible because our planned trajectories approximate the vehicle dynamics and thus allow maintaining maximal velocity---without losing control---when following them. 

\subsection{Collision}
\label{sec:collision}
\begin{figure}[h]
    \centering
        \begin{subfigure}{0.3\linewidth}
        \centering
        \includegraphics[width=\linewidth]{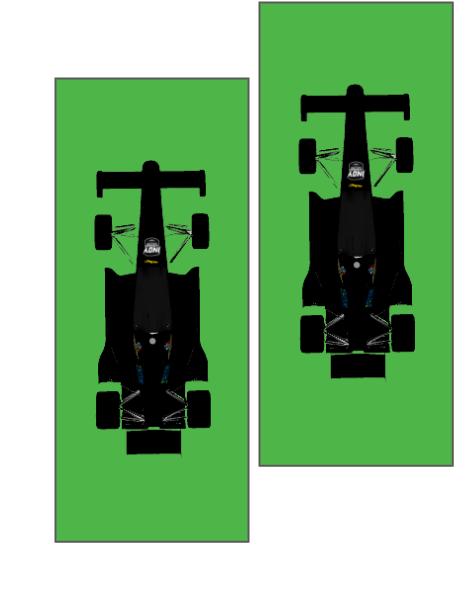}
        \caption{}
        \label{fig:saftey_bound_no_collision}
    \end{subfigure}\hspace{0.1\textwidth}%
    \begin{subfigure}{0.3\linewidth}
        \centering
        \includegraphics[width=\linewidth]{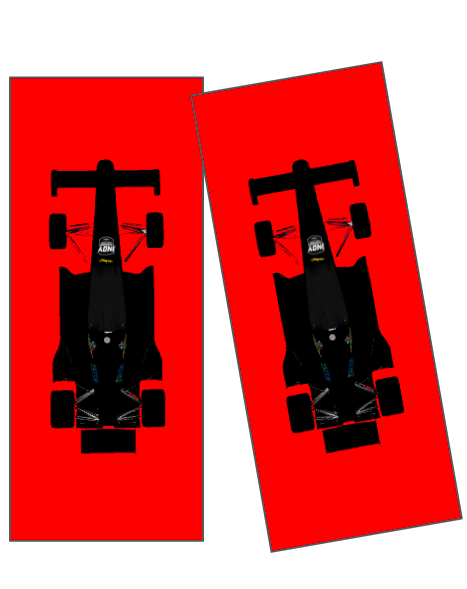}
        \caption{}
        \label{fig:saftey_bound_collision}
    \end{subfigure}
\caption{Rectangular safety bound. In (a) No overlap between safety bounds; (b) the safety bounds overlap.}
\label{fig:saftey_bound}
\end{figure}

To avoid collisions, we define a safety bound around the vehicle owing to the uncertainty inherent in our problem.
We attempt not only to avoid a collision with another vehicle but also to avoid any overlap between the safety bounds around both vehicles.
We use a rectangular safety bound, which best matches the vehicle's shape (as shown in Fig. \ref{fig:saftey_bound}). The vehicle's length is $5$ m, and its width is $2$ m. We defined the longitudinal safety bound as $0.3$ of the vehicle length, both front and rear, and the lateral safety distance as $0.5$ of the vehicle width, right and left. 
This creates a safety longitudinal distance of $6$ m between vehicles ($3$ m from each safety bound) and a lateral distance of $4$ m between vehicles. 
We note that the simpler circular safety bound is less appropriate for this case because of the length of the race car is more than twice its width.  

\begin{figure}[h]
    \centering
        \begin{subfigure}{0.5\linewidth}
        \centering
        \includegraphics[width=\linewidth]{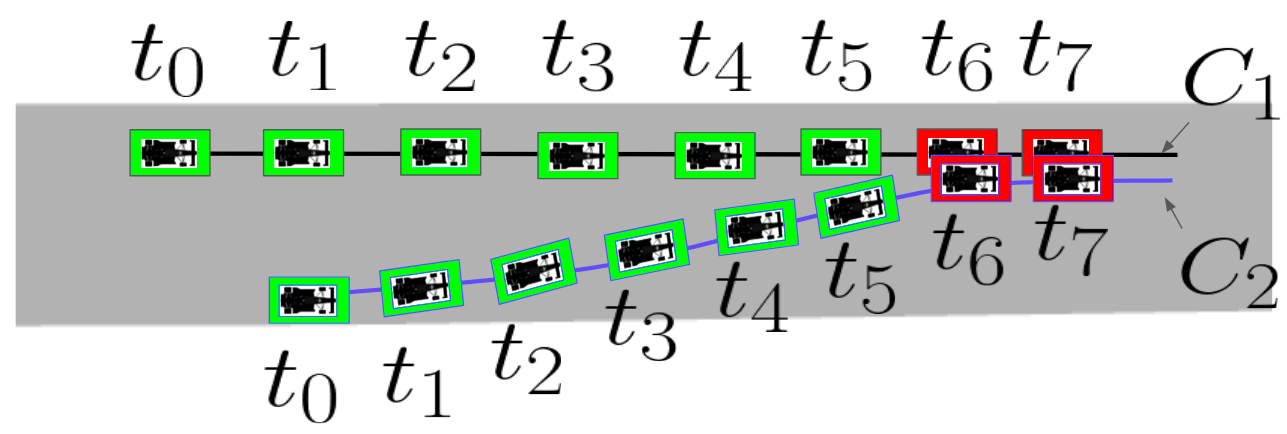}
        \caption{}
        \label{fig:future_collision2}
    \end{subfigure}%
    \begin{subfigure}{0.5\linewidth}
        \centering
        \includegraphics[width=\linewidth]{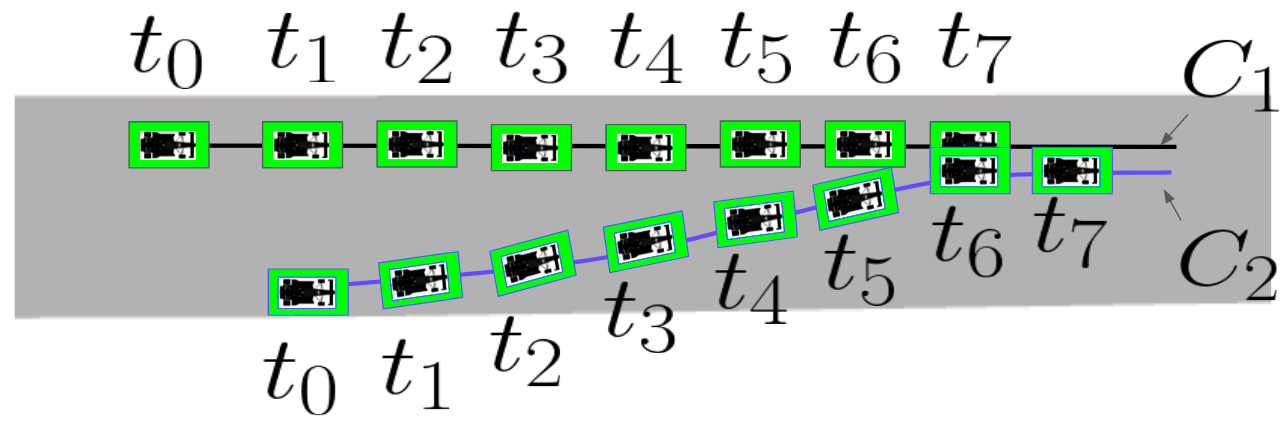}
        \caption{}
        \label{fig:future_collision1}
    \end{subfigure}
\caption{Two trajectories describing the location of each vehicle and its safety bounds. In (a), the vehicles collide at time $t_6$; in (b), the vehicles do not collide since the overlap of their safety bounds is at different times.}
\label{fig:future_collision}
\end{figure}

Two trajectories $C_1$ and $C_2$ collide  if
there exists some $t$ such that the safety bounds of both associated vehicles at time $t$ overlap. See Fig. \ref{fig:future_collision} for an illustration.

A maneuver candidate is considered \emph{free} if it does not collide with a predicted trajectory of any opponent vehicle. However, if an opponent vehicle is directly behind the ego-vehicle, and the ego-vehicle completely blocks it, our controller ignores it because changing the path to allow the opponent vehicle to overtake is clearly an uncompetitive behavior.

\subsection{Maneuver selection}
\begin{figure}[h]
    \centering
    \includegraphics[width=0.9\linewidth]{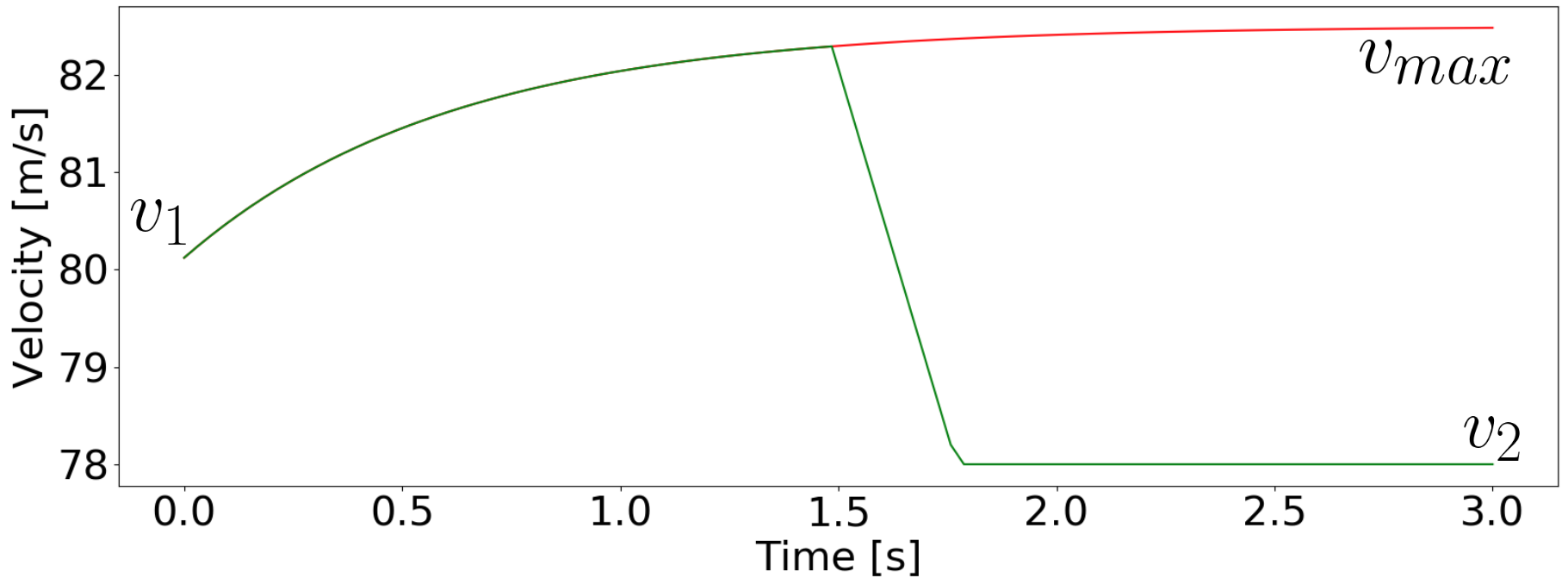}
    \caption{Updating a maneuver candidate to become free by decelerating towards a blocking vehicle that drives at a velocity of $v_2$ (green), instead of accelerating to the maximal velocity, $v_{\textrm{max}}$ (red), and causing a collision.}
    \label{fig:updated_velocity}
\end{figure}

Our planner considers all free maneuver candidates. Colliding candidates that assume maximal velocity are updated by reducing the speed along them to avoid collision, as shown in
Fig. \ref{fig:updated_velocity}.

If more than one maneuver is free, 
each maneuver is given a cost that is the sum of three criteria:  travel time, denoted $\mathcal{T}(C)$,  nearness to the optimal race line, denoted $\mathcal{N}(C)$, and continuity, denoted $\mathcal{K}(C)$, as detailed below. This produces the cost function $\text{Cost}(C)$ for  each free maneuver:
\begin{equation}
    \text{Cost}(C)= \mathcal{T}(C)-\mathcal{N}(C)-\mathcal{K}(C);  C \in \boldsymbol{M}.
\end{equation}

\subsubsection{Travel time}
$\mathcal{T}(C)$  represents the estimated travel time for maneuver $C$.  It is computed    
by integrating the maneuver's velocities along $C$.

\subsubsection{Nearness to the optimal race line}
The search for the fastest maneuver with a limited time horizon is local by nature.  
A global search is ineffective because of the unpredictable behavior of the opponent vehicles.     
A sensible compromise is to attempt to merge with 
the precomputed race line, which is globally optimal, wherever possible.  
The function $\mathcal{N}(C)$ thus equals  $R_\text{opt}$ if $C$ is the closest to the optimal race line among the free candidates and $0$ otherwise.

\subsubsection{Continuity}
To avoid frequent oscillations between 
maneuver candidates of similar optimality, 
it is preferred, when possible, to maintain the same maneuver unless another maneuver is conspicuously better. This is done by rewarding the maneuver that is similar to the current maneuver.   For maneuver $C$, $\mathcal{K}(C)$ equals $R_{k}$ if $C$ is the same maneuver as the maneuver that the vehicle currently drives on and $0$ otherwise.
Clearly, switching to a new maneuver is less desired if a switch has just occurred, but once some time has elapsed since the last switch, the controller should be more lenient towards another switch. This is accomplished by 
decaying $R_{k}$  linearly at the rate $R_{d}$, as long as the same maneuver has been followed.  

\begin{figure}[h]
    \centering
    \begin{subfigure}{\linewidth}
        \centering
        \includegraphics[width=\linewidth]{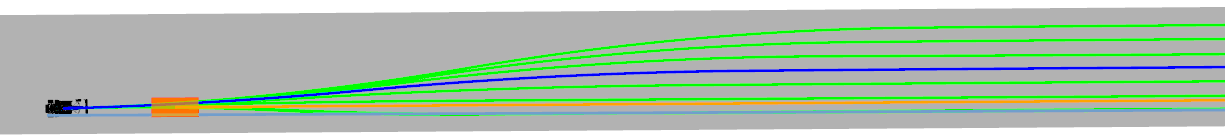}
        \caption{}
        \label{fig:min_time_overtake_short}
    \end{subfigure}
    \begin{subfigure}{\linewidth}
        \centering
        \includegraphics[width=\linewidth]{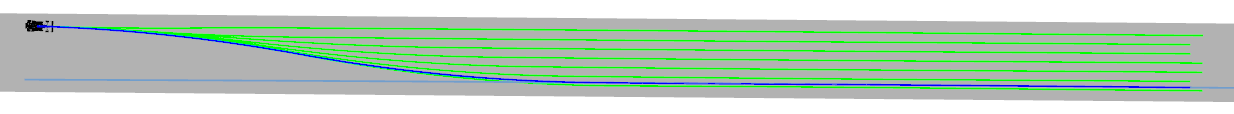}
        \caption{}
        \label{fig:from_left_to_optimal}
    \end{subfigure} 
    \begin{subfigure}{\linewidth}
        \centering
        \includegraphics[width=\linewidth]{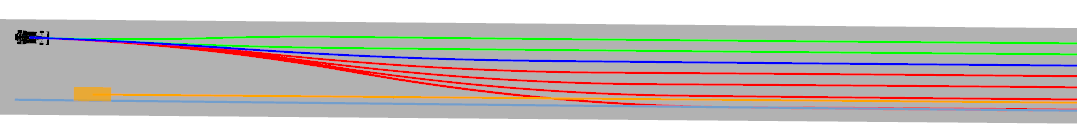}
        \caption{}
        \label{fig:closest_to_optimal}
    \end{subfigure} 
    \begin{subfigure}{\linewidth}
        \centering
        \includegraphics[width=\linewidth]{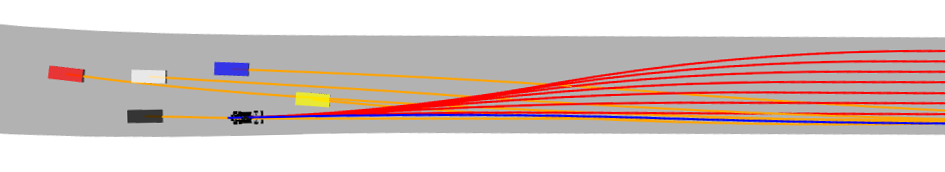}
        \caption{}
        \label{fig:all_blocked}
    \end{subfigure} 
\caption{Maneuver selection examples, green: free maneuvers, red: blocked maneuvers, blue: chosen maneuver. (a) The planner selects a minimum-time lateral-shift maneuver to avoid slowing down because of the blocking vehicle. (b) The planner prefers the longer maneuver because it returns to the optimal race line (shown in light blue). (c) The planner selects a free maneuver that is the closest to the optimal race line. (d) None of the maneuver candidates are free; therefore, the planner selects the safest maneuver.}
\label{fig:trajectory_choice}
\end{figure}

\subsection{Behavior when no maneuver is free}
In dynamic environments, such as a multi-vehicle race, there might occur situations when no free maneuver exists.     
In such sitatuations, the optimality criteria are irrelevant, and then the only selection criterion is safety. The planner then determines the  imminent collision and selects the maneuver that is as far as possible from that collision (see an example in Fig. \ref{fig:all_blocked}). 

\section{Control}
\label{sec:control}
The control module outputs throttle, brake, and steering commands that drive the vehicle as close as possible to the selected maneuver. %
\subsection{Lateral control}
\begin{figure}[t]
    \centering
    \includegraphics[width=0.5\linewidth]{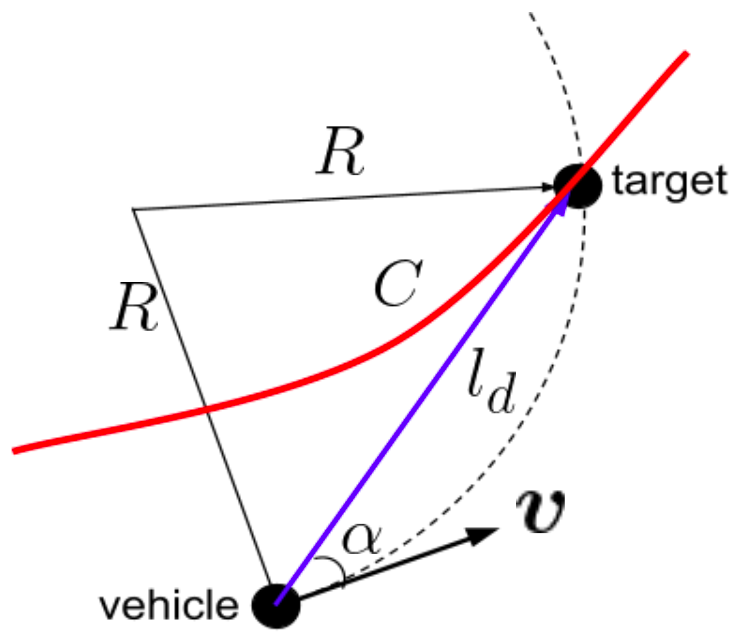}
    \caption{Pure pursuit geometry. The red line represents the reference trajectory $C$; the algorithm finds a curve with a constant curvature $1/R$ that will lead the vehicle to a target on $C$ that is at a distance $l_d$ from the vehicle.}
    \label{fig:pure_pursuit}
\end{figure}

The pure pursuit algorithm \cite{pure_pursuit} is used to compute the desired angular velocity based on the current ego-vehicle's state and the selected maneuver, which is mapped to the Cartesian coordinates. The pure pursuit algorithm pursues a target on the selected maneuver $C$. Let $\boldsymbol{v}$ be the vector representing the ego-vehicle's velocity. The distance to the target is defined to be proportional to the vehicle's speed $v = |\boldsymbol{v}|$,  that is, $l_d = k_tv$, where $k_t$ is a predefined constant. The angle between the velocity vector $\boldsymbol{v}$ to the vehicle-target vector is denoted as $\alpha$ (see Fig. \ref{fig:pure_pursuit}).
The desired angular velocity of the vehicle, $\omega_d$ is
$2v\sin{\alpha}/{l_d}$.
The desired angular velocity $\omega_d$ is used as a reference for a proportional angular velocity controller that computes the steering command:
$\delta = k_{\omega}(\omega_d - \omega)$
where $\omega$ is the current angular velocity of the vehicle and $k_{\omega}$ is a proportional gain.

\subsection{Longitudinal control} 
\begin{figure}[h]
    \centering
    \includegraphics[width=0.7\linewidth]{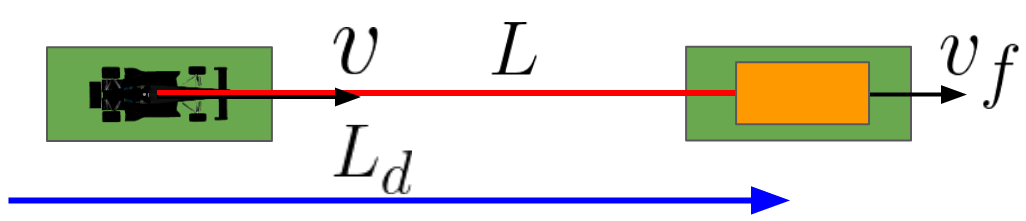}
    \caption{The ego-vehicle follows another vehicle. The distance between the ego-vehicle and the vehicle ahead, $L$, is less than the predefined following distance, $L_d$. Therefore, the ego-vehicle will slow down to increase $L$. }
    \label{fig:following}
\end{figure}

The desired speed $v_d$ is provided by the selected maneuver, which is typically used as a reference for the longitudinal controller.
However, when closely following a vehicle, we modify the desired speed to:
$v_d = v_f - k_f(L_d - L)$,
where $v_f$ is the speed of the leading vehicle, $L$ is the distance to the leading vehicle, $L_d$ is the desired distance to maintain, and $k_f$ is a proportional gain. 
This modification allows for smooth driving and maintaining a constant distance from the leading vehicle. Figure \ref{fig:following} illustrates the vehicle-following scenario.
Finally, the throttle and brake command $u$ is computed by a proportional speed controller:
$u = k_v(v_d - v)$
where $k_{v}$ is a proportional gain.

We note that the vehicle was operated most of the time at high speeds where the vehicle has low acceleration capabilities because of the high aerodynamic drag at these speeds. Therefore, we could assume that the longitudinal dynamics are linear and tune the longitudinal controller for the best performance at those speeds range.

\section{Experiments}
\label{sec:Experiments}
\subsection{Simulation environment}
\begin{figure}[h]
    \centering
        \includegraphics[width=0.7\linewidth]{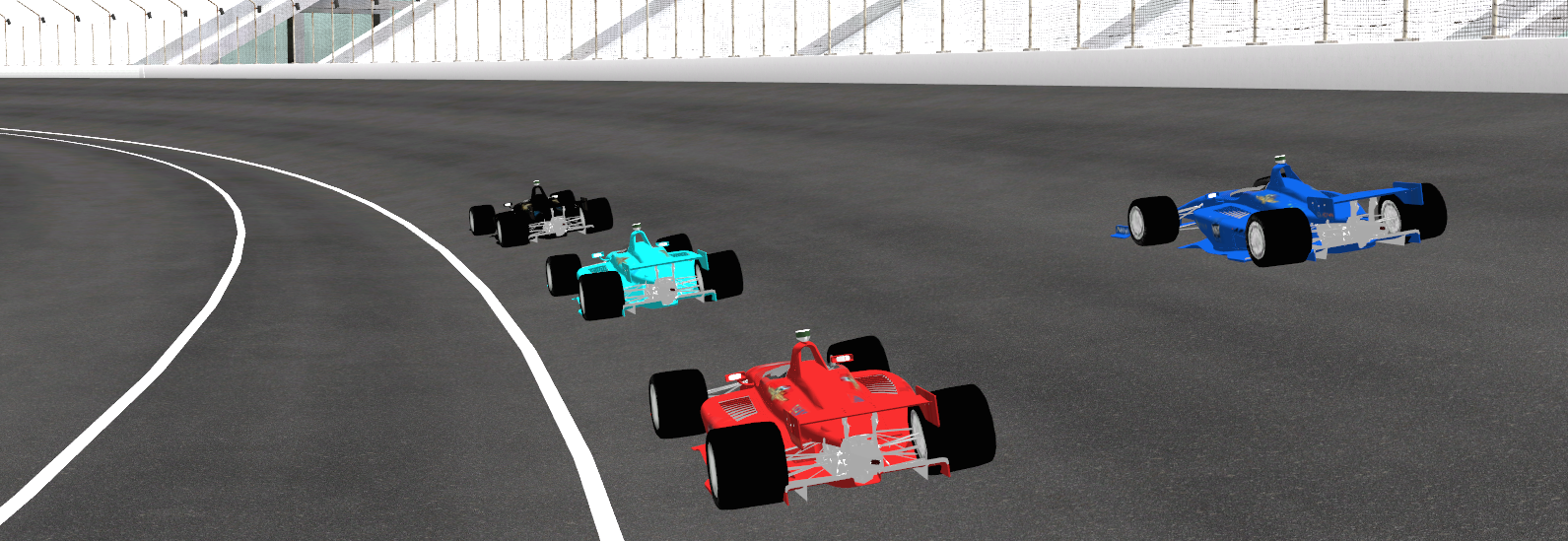} %
    \caption{The VRXPERIENCE simulator used for the simulation race.} 
    \label{fig:simulation_close_look}
\end{figure}

The simulation platform used by all teams was the Ansys VRXPERIENCE simulator \cite{ansys_sim}, which simulates the AV-21 dynamics and the IMS track (see Fig. \ref{fig:simulation_close_look}).
The VRXPERIENCE simulator enables multi-vehicle head-to-head racing, in which every vehicle is controlled by a separate controller.
The sensors are simulated at $25$ Hz and the ego-vehicle's state at $100$ Hz, in simulator time. 
Each controller receives the ego-vehicle's state and the sensor data from the simulator and sends back the throttle, brake, and steering commands. 

We use the pre-processed cameras and radars data from the simulator to obtain the position and linear and angular velocity of all vehicles in the sensors range. Our controller is based on Autoware.auto \cite{Autoware}, which is an open-source autonomous driving framework that uses ROS2 as middleware. The computation time of each cycle at our algorithm is $20$ milliseconds, on average, on an Intel Core i7 $2.90$ GHz CPU, which enables our algorithm to run in real time. 
A video demonstrating our controller is available in \cite{summary_video}.   

\subsection{Simulation results} 
\begin{figure}[h]%

    \centering
    \begin{subfigure}{\linewidth}
        \centering
        \includegraphics[width=\linewidth]{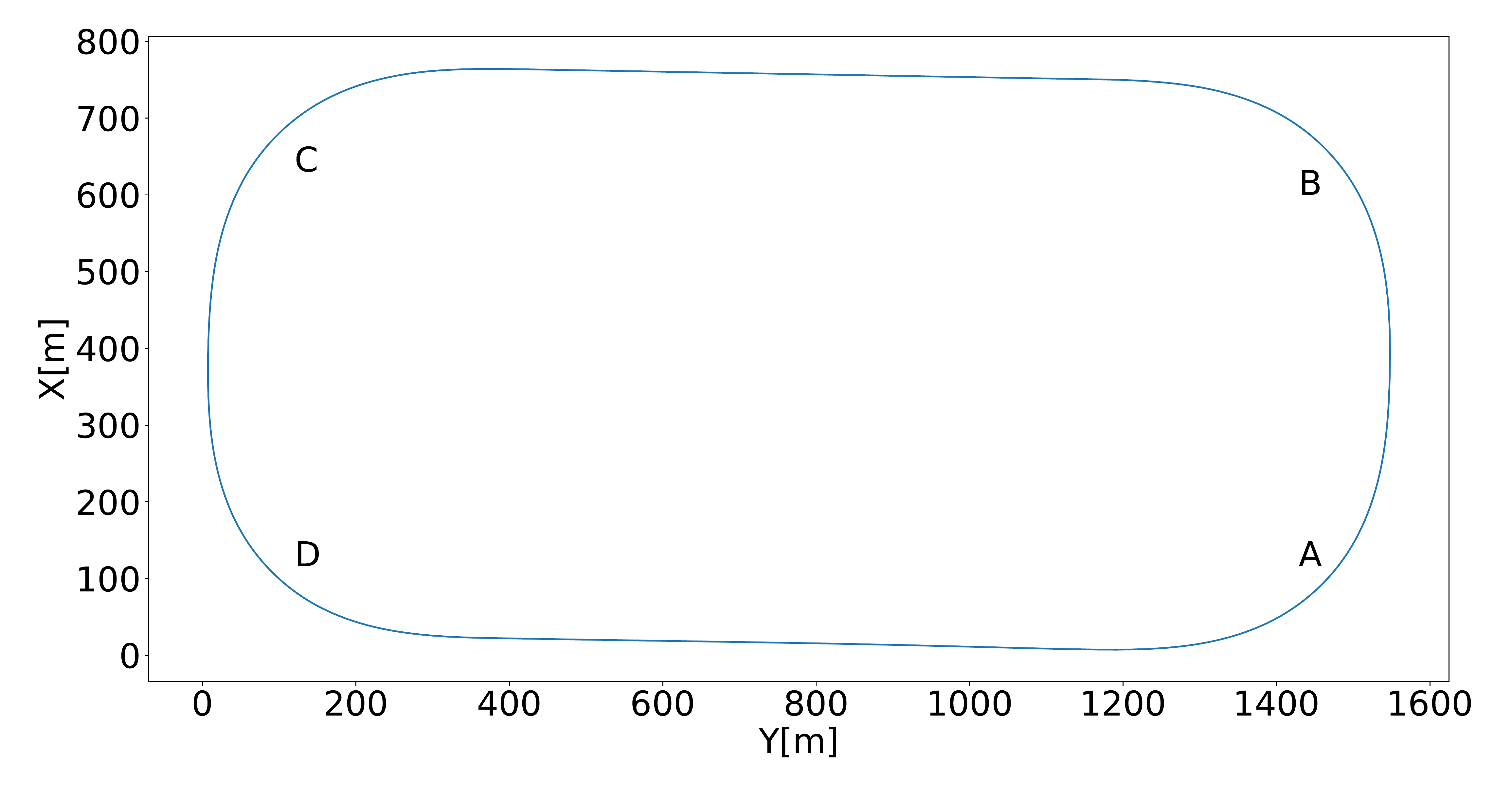}
        \caption{}
        \label{fig:trajectory}
    \end{subfigure} 
    
    \begin{subfigure}{\linewidth}
        \centering
        \includegraphics[width=\linewidth]{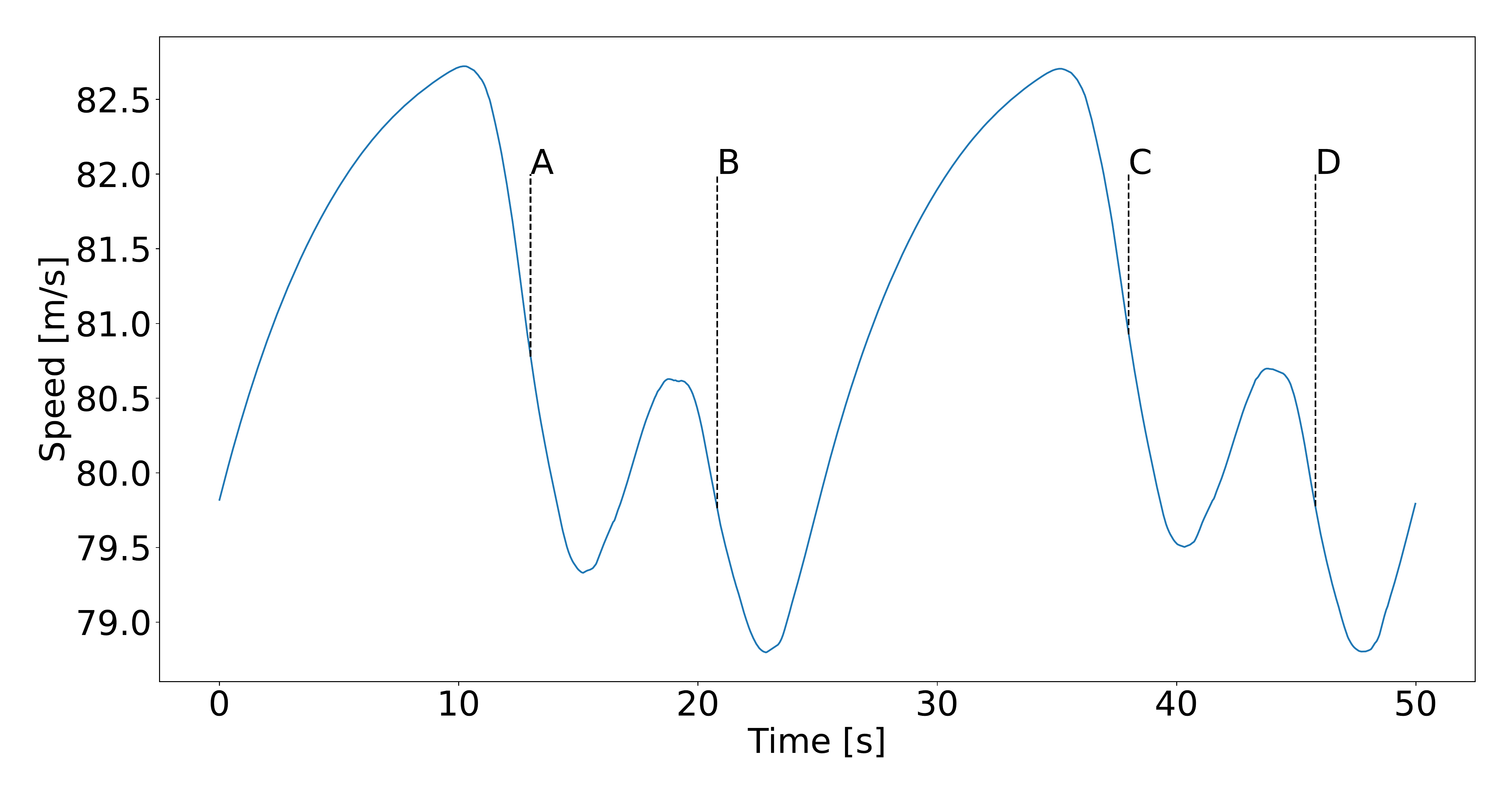}
        \caption{}
        \label{fig:solo_lap_vel}
    \end{subfigure} 
    
     \begin{subfigure}{\linewidth}
        \centering
        \includegraphics[width=\linewidth]{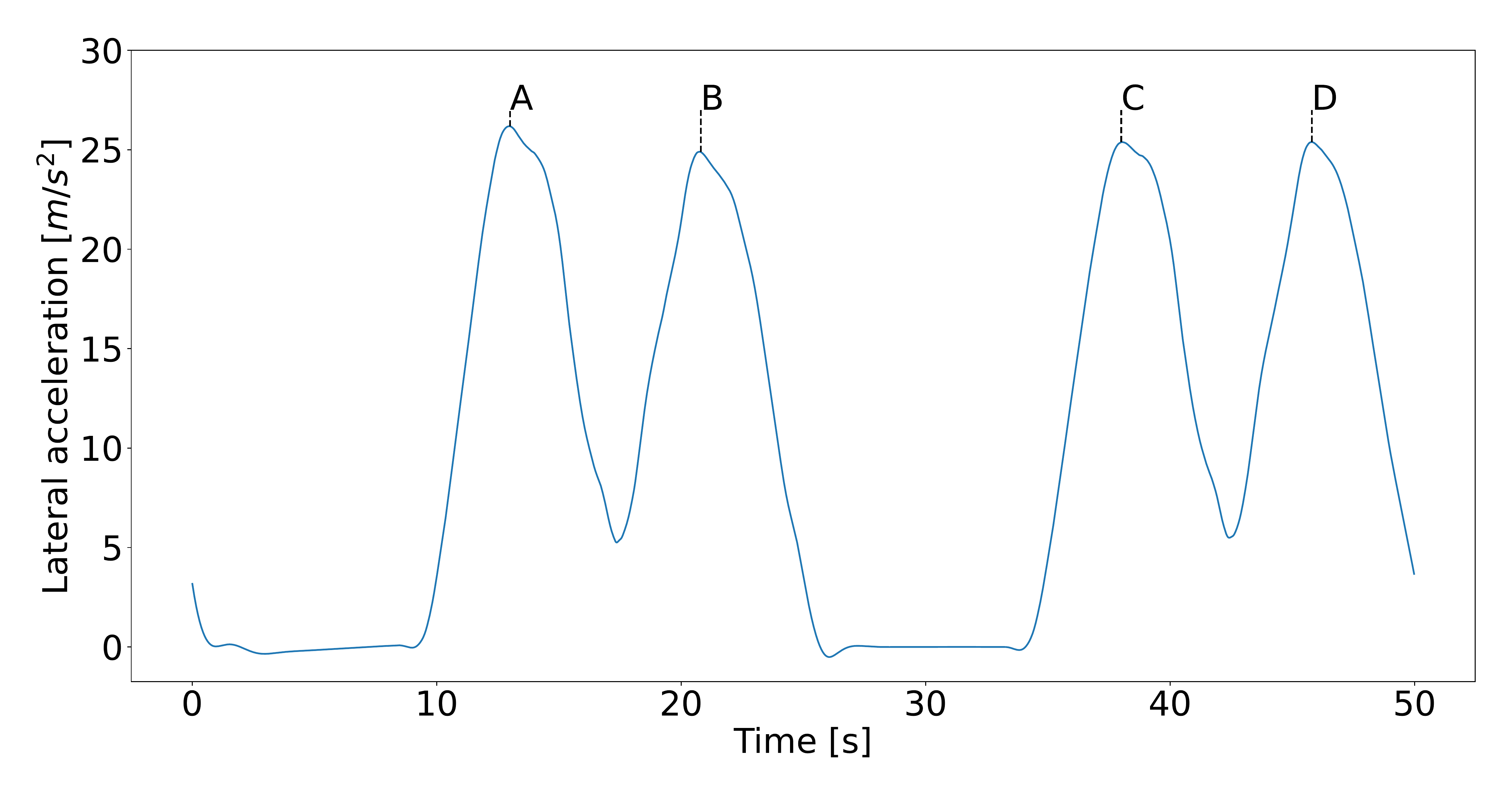}
        \caption{}
        \label{fig:solo_lap_acc_y}
    \end{subfigure} 
    \caption{(a) Path, (b) velocity profile, and (c) lateral acceleration during the solo lap. The vehicle slows down from a speed of $82.72$ m/s to $78.79$ m/s in the corners, labeled A-D. The lateral acceleration in the corners exceeds $2.5$ g due to the high down-force.}%
    \label{fig:solo_lap}%
\end{figure}

We first tested the solo lap performance of our controller, that is, driving along the track without other vehicles. The lap time was $50.0$ seconds, and the vehicle had an average speed of $80.83$ m/s.
The vehicle ran at full throttle along the entire lap. The speed and the lateral accelerations during a single lap are shown in Fig. \ref{fig:solo_lap}. As depicted by the figure, the vehicle's speed dropped slightly at the corners from a top speed of $82.72$ m/s to $78.79$ m/s. That is owing to the increased tire slip, which detracts from the longitudinal tire force during the turns. Furthermore, the lateral acceleration during the turns was over $2.5$ g, which is way above what is expected of a regular passenger car, reflecting the high down-force generated by the race car due to its high speed and aerodynamic properties.
\newcommand{\xf}{3.5cm}

We then tested our controller in a multi-vehicle scenario, running the same controller on  $6$ vehicles over $30$ laps. Every controller instance operated independently of the other controllers and received information only from its own vehicle sensors. During the entire test, all vehicles drove safely while respecting the race rules; a collision or loss of control never occurred.
Figure \ref{fig:6_ego} shows two typical snapshots from the planner of the black vehicle. Notably, all vehicles are very close to the leading vehicle (approximately $40$ m) while moving at very high speeds (over 80 m/s), resulting in approximately $0.5$ s separating the leading vehicle from the last.  
A recording of this test, which took $28$ minutes, is available at \cite{multi_vehicle_test} and the raw data at \cite{data_log}. 

\newcommand{\xe}{3.5cm}
\begin{figure}[h]%
    \centering
    \subfloat[\centering ]{{\includegraphics[height=\xe ]{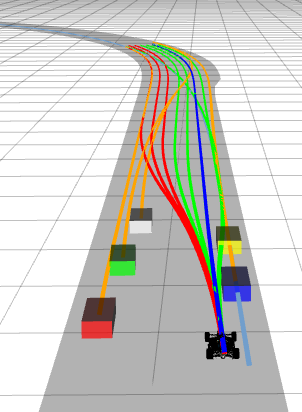} }}%
    \subfloat[\centering ]{{\includegraphics[height=\xe ]{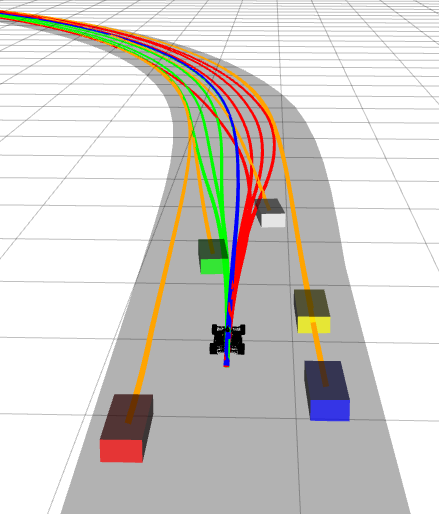} }}%
    \caption{Two snapshots of trajectory planning a scenario with 5 opponent vehicles. Predicted opponent trajectories (orange);  free maneuvers (green); colliding maneuvers (red); selected maneuver (blue). (a) The ego vehicle drives on a straight. The colliding maneuvers are close to the inner boundary of the track. The selected maneuver continues on a straight line. (b) The vehicles are approaching a corner. The colliding maneuvers are close to the outer side boundary of the track. The selected maneuver overtakes the green vehicle from the right.}%
    \label{fig:6_ego}%
\end{figure}

\begin{figure}[h]
    \centering
    \includegraphics[width=1.0\linewidth]{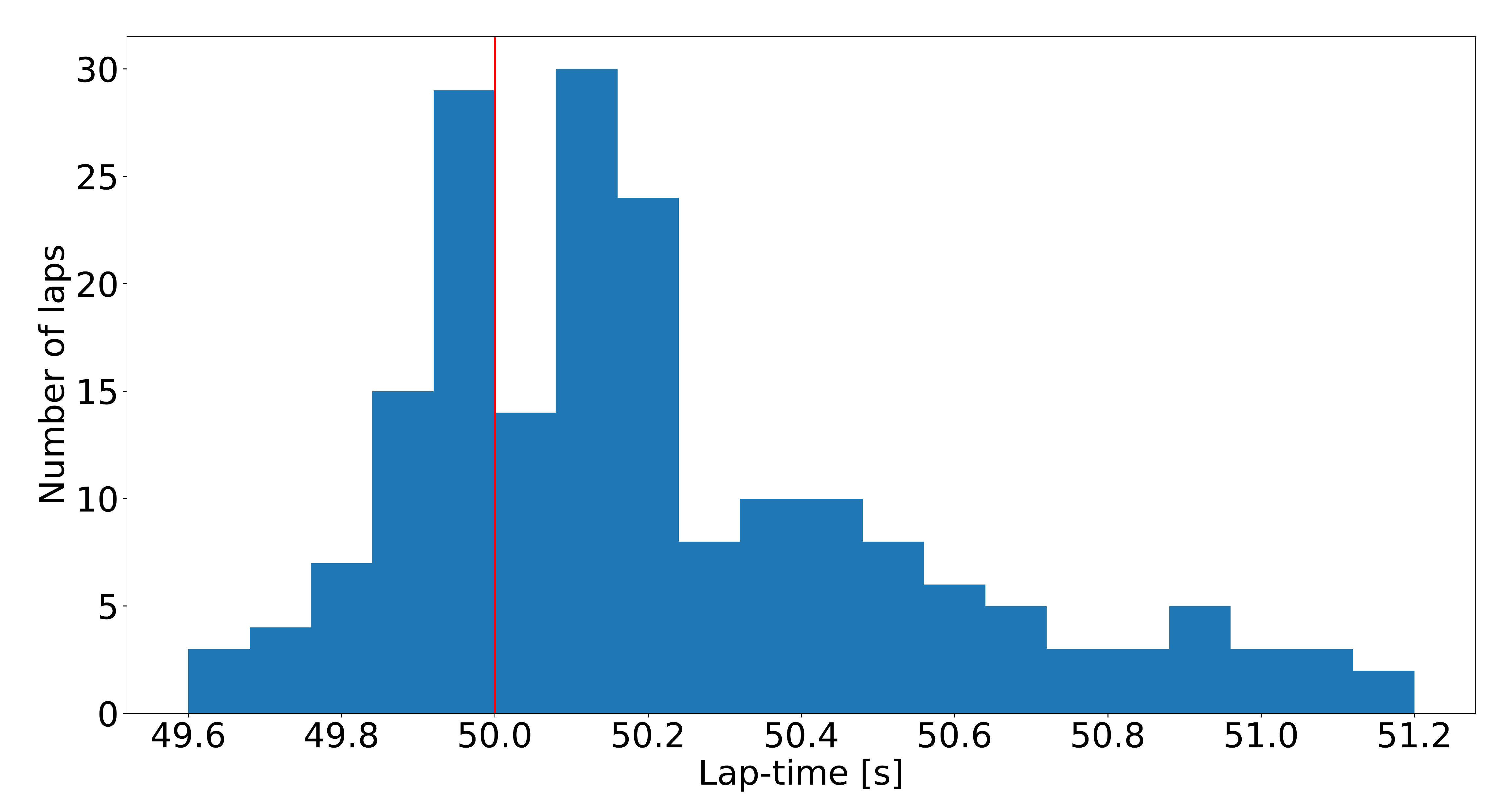}
    \caption{Lap times distribution of all $6$ vehicles over  $30$ laps. The red line represents the solo-lap time. }
    \label{fig:lap_time_distribution}
\end{figure}

The lap times in the multi-vehicle test ranged from $49.84$ to $51.24$ seconds, a difference of less than $3\%$ between the extremes. 
The average lap time was $50.25$ seconds, which is very close to the solo lap time of $50.0$ seconds. The lap-time distribution is shown in Fig. \ref{fig:lap_time_distribution}.  Interestingly, some of the individual laps were faster than the solo-lap. This is owing to the higher speeds that individual vehicles can achieve while taking advantage of the slipstream of the leading vehicles. 

\begin{figure}[h]
    \centering
    \includegraphics[width=1.0\linewidth]{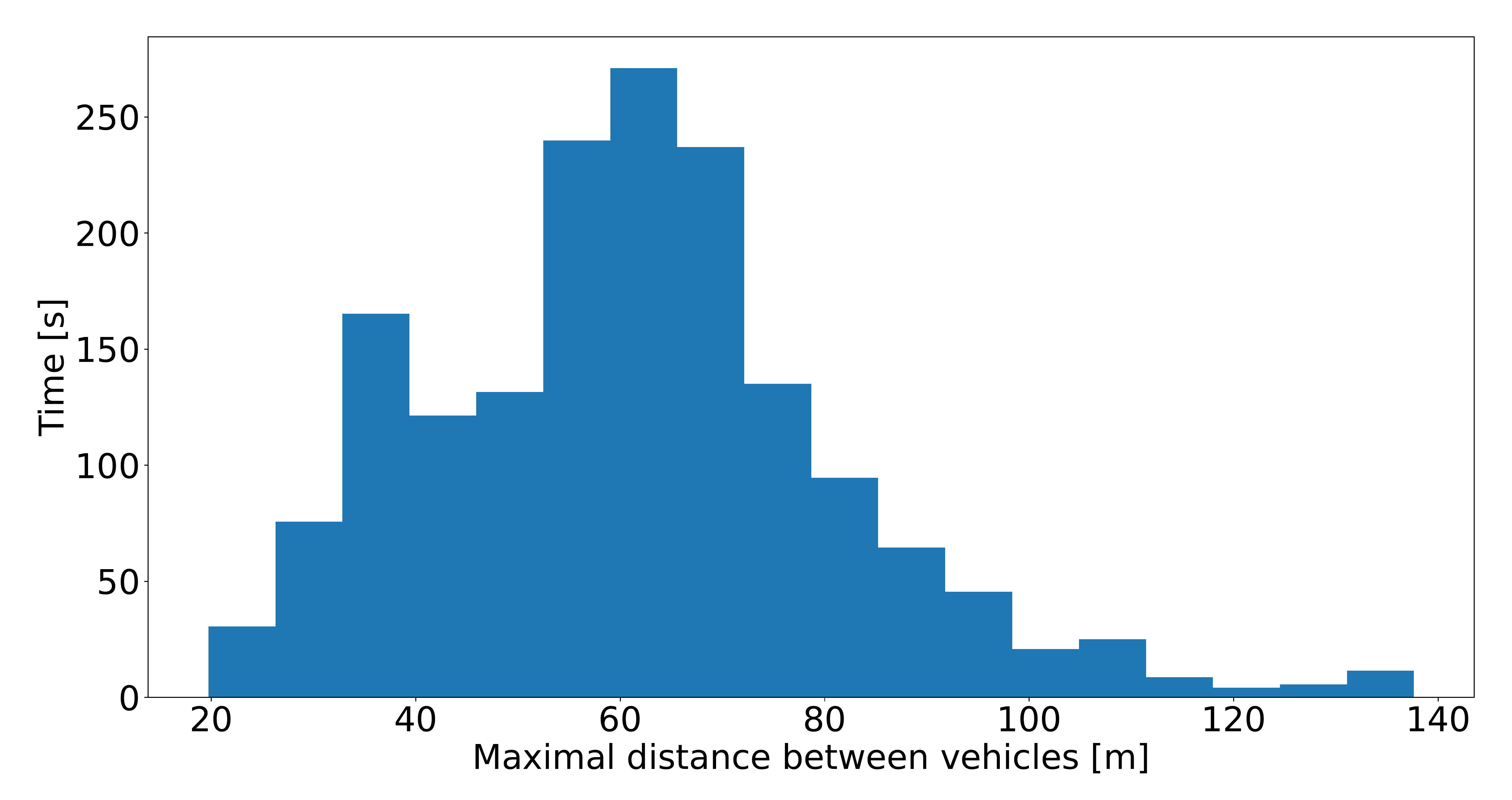}
    \caption{Distribution of the longitudinal distance between first and last vehicle. }
    \label{fig:max_distance}
\end{figure}

Figure \ref{fig:max_distance} shows the distribution of the longitudinal distance between the first and last vehicles over the $30$ laps. The average distance was $64.7$ m, which indicates a very tight race.

\begin{figure}[h]
    \centering
    \includegraphics[width=\linewidth]{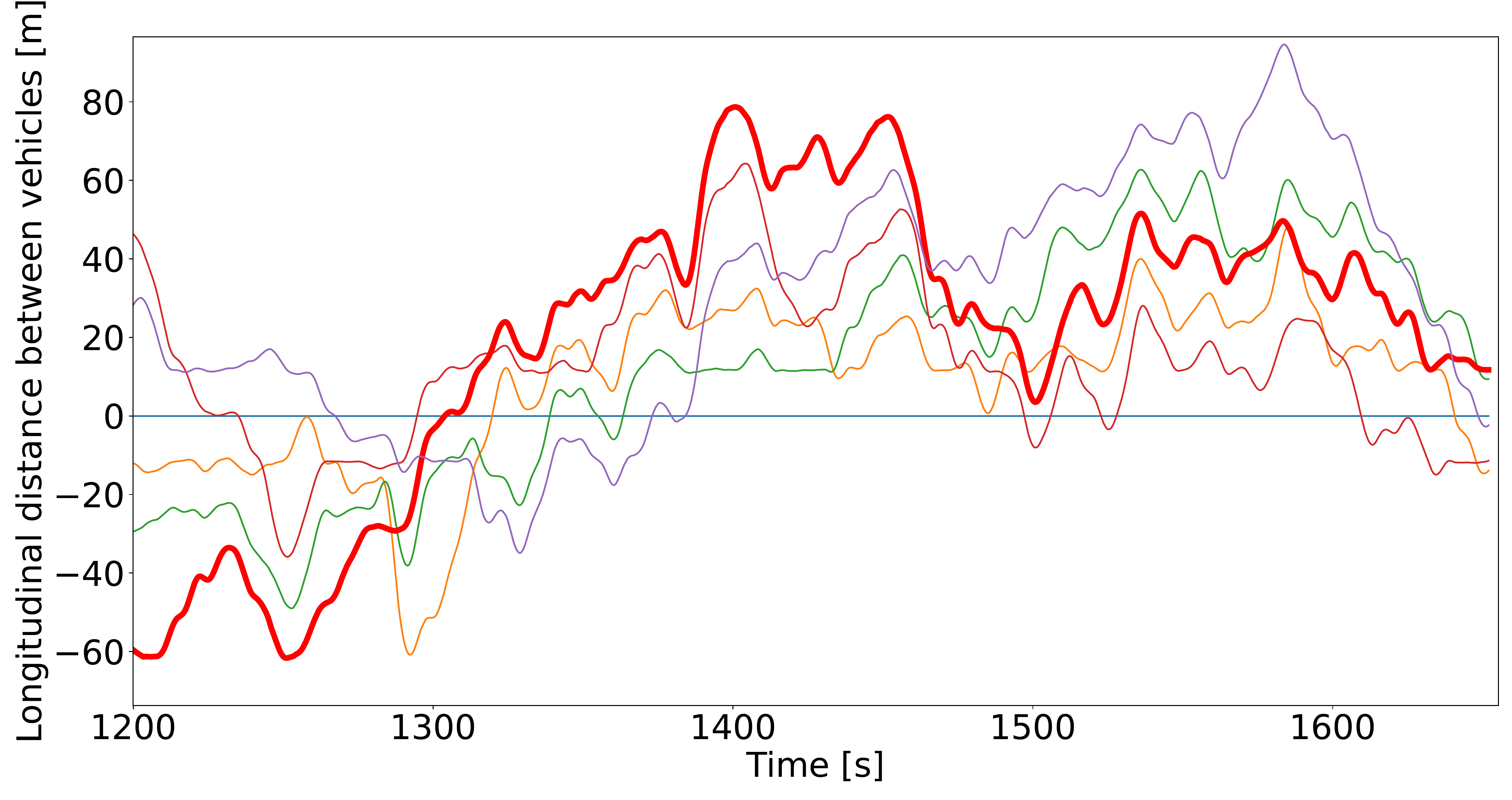}
    \caption{The longitudinal distance from each vehicle (each represented by a line with a different color) relative to the first vehicle over time. It shows a time frame of the last $10$ laps of the multi-vehicle racing test. Intersections between the lines represent overtakes between vehicles, which indicates that the vehicles repeatedly overtake each other. The marked red line shows that this vehicle overtook all vehicles and finished first.}
    \label{fig:longitudinal_distances}
\end{figure}

Another indication of the race tightness and the competitive behavior of all vehicles is shown in Fig. \ref{fig:longitudinal_distances}. It shows the longitudinal distance from all vehicles to the first vehicle during the last $10$ laps. Remarkably, the vehicle starting far behind all other vehicles  (represented by the red line) crossed the finish line ahead of the other vehicles.
The last vehicle crossed the finish line only $0.32$ seconds after the first vehicle.
Note that, as mentioned in \ref{sec:Aerodynamics_forces}, the vehicles can overtake each other, although having identical dynamics, by exploiting the slipstream generated by the leading vehicle.

Figure \ref{fig:takeover_trail} shows a sequence of snapshots of one overtaking maneuver with three competing vehicles. As shown, the black vehicle, starting second in Fig. \ref{fig:takeover_trail}b, gains a slight advantage over the blue vehicle at every corner until it successfully overtakes the blue vehicle, as shown in Fig. \ref{fig:takeover_trail}g. 
This overtaking maneuver took approximately $40$ seconds and $3$ km to complete.    
We note that explicitly planning such a maneuver requires a planning horizon of a few kilometers, which is computationally expensive. Nevertheless, our planner executes such maneuvers by repetitive local planning over a horizon of $3$ seconds. Although some of our controller's attempts to overtake other vehicles may fail, it does not compromise the vehicle's safety.

\newcommand{\xc}{4.3cm}
\newcommand{\xd}{3.9cm}
\begin{figure}[h]%
    \centering
     \begin{subfigure}{\linewidth}
        \centering
        \includegraphics[width=\linewidth]{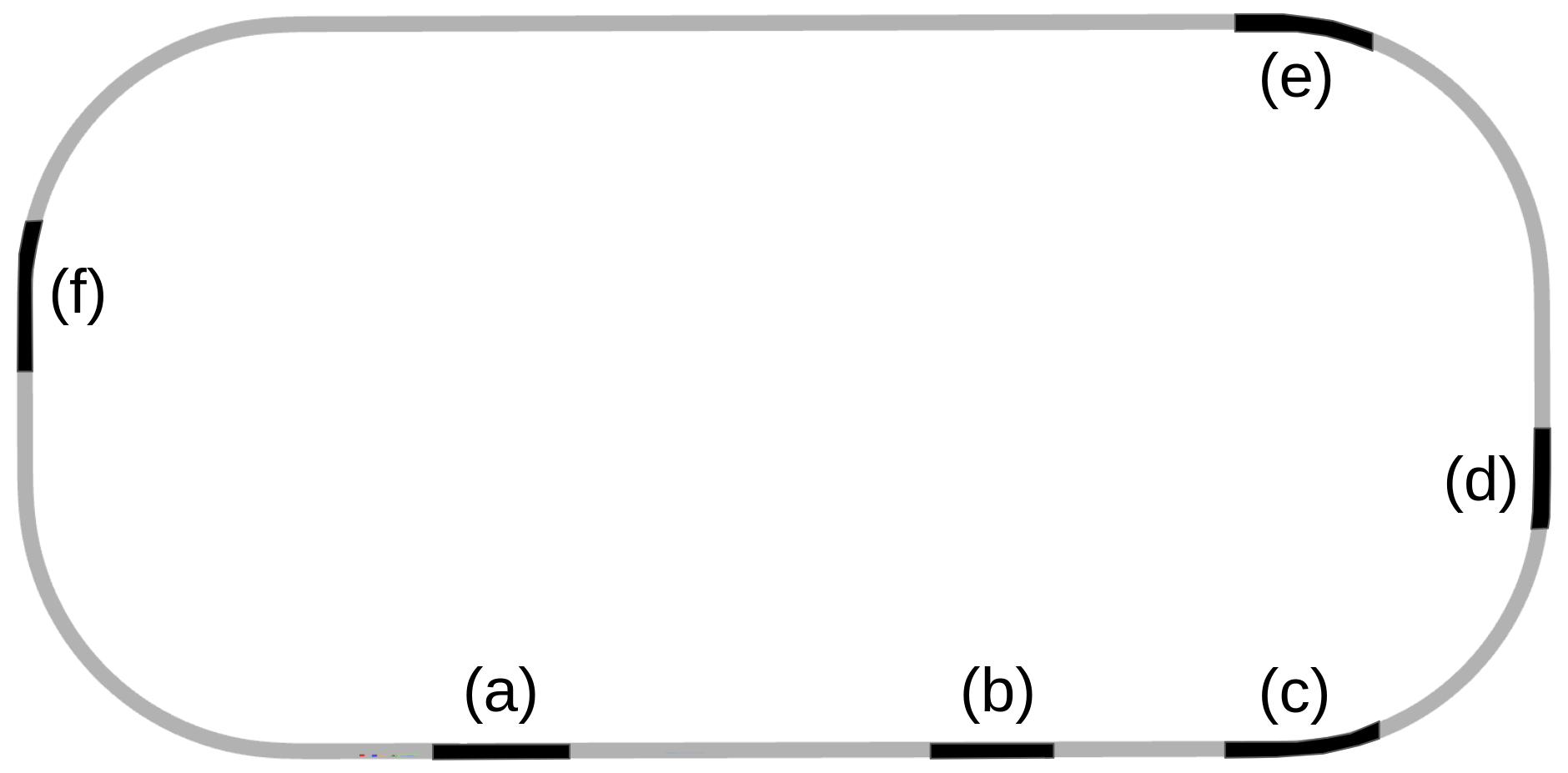}
        \label{fig:on_track}
    \end{subfigure} 
    
    \subfloat[\centering ]{{\includegraphics[width=\xc]{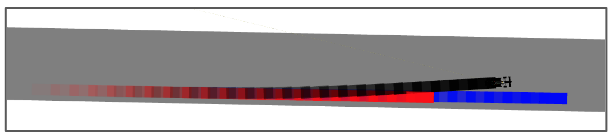} }}%
    \subfloat[\centering ]{{\includegraphics[width=\xd]{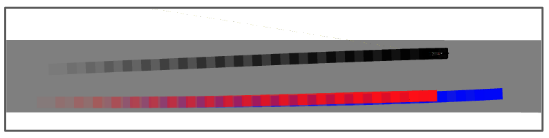} }}%
    
    \subfloat[\centering ]{{\includegraphics[width=\xc]{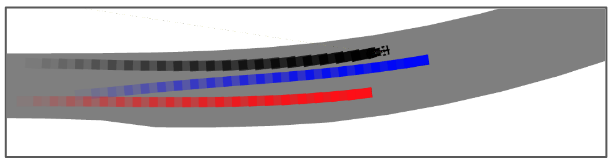} }}%
    \subfloat[\centering ]{{\includegraphics[width=\xd]{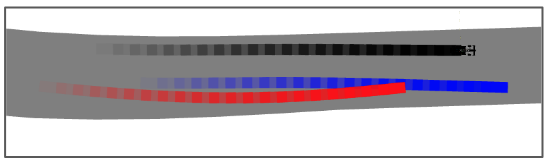} }}%
    
    \subfloat[\centering ]{{\includegraphics[width=\xc]{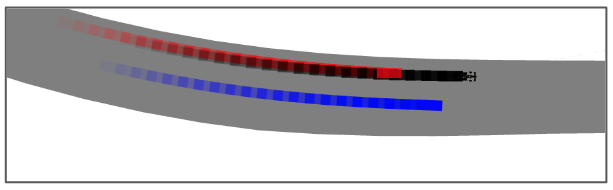} }}%
    \subfloat[\centering ]{{\includegraphics[width=\xd]{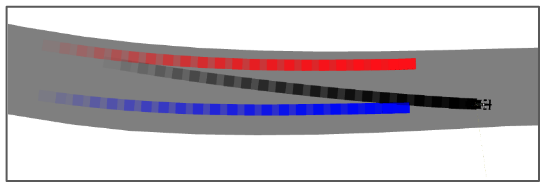} }}%

    \caption{A sequence of snapshots of an overtaking maneuver---the black vehicle overtakes the blue vehicle. Each snapshot lasts $0.8$ seconds. The track with marked snapshot segments is depicted on the top. (a) The black vehicle initiates an overtake. (b) The black vehicle drives parallel to the blue and red vehicles. (c) The black vehicle prevents the blue vehicle from continuing on the optimal race line, i.e., reaching the apex. (d) The black vehicle gains a slight advantage at the corner. (e) At the next corner, the blue vehicle is overtaken because it is forced to stay on the outside of the corner. (f) The black vehicle returns to the optimal race line.}%
    \label{fig:takeover_trail}%
\end{figure}

\subsection{The IAC simulation race}
Sixteen teams reached the simulation race. All teams first competed on one solo lap, with an initial speed of $100$ km/h (rolling start), to determine the vehicles' order for the multi-vehicle race.  
Our solo lap time of $51.968$ seconds  placed us in the 7\textsuperscript{th} place;  the first place finished at $51.848$, 
only $0.12$ seconds ($0.23\%$) ahead of us. 
The average time of all teams was $53.041$ seconds.

Only $10$ teams passed the multi-vehicle safety tests and were qualified to proceed to the semi-final, which
consisted of a multi-car, $10$-lap competition. The $10$ teams were split into two heats, $5$ vehicles on each. Based on our solo lap time, we were placed in the 3\textsuperscript{rd}  place in our heat. We finished the semi-final  in the 3\textsuperscript{rd} place, at $505.428$ seconds, only $0.44$ seconds ($0.0871\%$) after the winner.
The average time for this heat was $506.711$ seconds. 
In the second heat, three of the five teams lost control or crashed. That left seven teams for the final race.

We started the final race in 5\textsuperscript{th} place, and our vehicle overtook two vehicles over the first lap. 
Throughout the race, our controller demonstrated collision-free and competitive driving capabilities and was able to keep the 3\textsuperscript{rd} place a significant part of the time. With three laps to go, another vehicle entered our safety zone from the right, which triggered a collision avoidance action. Being on the far left track, with no room to maneuver, the collision was avoided by stepping on the brakes. Driving at that moment along the turn caused our vehicle to spin off the track, which placed us at the 6\textsuperscript{th} place in the finals. 
Four vehicles completed the race without being responsible for a collision.
The recording of the simulation race is available in \cite{race_recording}.

\subsection{Remarks}
One of the interesting results accomplished by the racing controller presented here is its ability to balance competitiveness and safety.  This was demonstrated by driving $6$ vehicles for $30$ laps each, which accounts for  $180$ laps over $720$ km, without any collision.  This resonates with our collision-free run in the IAC simulation race over a total of $17$ laps.  

This notable result did not diminish the racing controller's competitiveness, as demonstrated by the 6-vehicle race, where the average lap time of all vehicles was only $0.25$ seconds greater than the optimal solo-lap time.  This implies that the vehicles were running in close proximity to each other, thus demonstrating the great challenge of high-speed racing.   

It is interesting to note that despite all vehicles being driven by the same controller, they continuously overtook each other by exploiting the slipstream from the leading vehicle, as was depicted in Fig. \ref{fig:longitudinal_distances}.

\section{Conclusions} 
This paper describes a competitive racing controller for an autonomous racing car developed in the context of the Indy Autonomous Challenge simulation race. Its development was guided by an attempt to strike a balance between competitiveness and safety. To this end, the controller attempts to avoid collisions, even those that the race rules placed the responsibility on the opponent vehicle to avoid.   

The online planner generates a set of dynamically feasible maneuver candidates using a point mass model.
Of those candidates, a maneuver is selected that is collision-free, minimizes travel time along the track, and maximizes proximity to the race line. It is then tracked by a pure-pursuit controller. The speed is controlled by a speed controller that follows the velocity profile along the trajectory and regulates the speed to avoid collision with neighboring vehicles.   

Our controller demonstrated competitive and safe driving in a test run with $6$ vehicles, all driven by the same controller, and in the IAC simulation race.  
Our vehicle finished 3\textsuperscript{rd} in the semi-finals with only $0.44$ seconds behind the winner, and 
maintained 3\textsuperscript{rd} place for a significant part of the final race.  %
It demonstrated responsible driving, yet competitive, and was not involved in any collision during the entire race.  
It is important to note that very few vehicles were not involved in any collision with any other vehicle. 
Clearly, while the challenge of safe and competitive driving is still unresolved, the Indy Autonomous Challenge competition brought us closer to driving autonomously under extreme conditions.  

\section{Acknowledgments} 
The authors wish to thank the Indy Autonomous Challenge organizers and the Ansys Team for providing 
a unique opportunity to test our algorithms in a high fidelity simulation and compare the performance of various methods on a common ground. The first author acknowledges the generous support of Ariel University throughout his graduate studies.

\bibliographystyle{IEEEtran.bst}
\bibliography{indy_paper}

\begin{IEEEbiography}[{\includegraphics[width=1in,height=1.25in,clip,keepaspectratio]{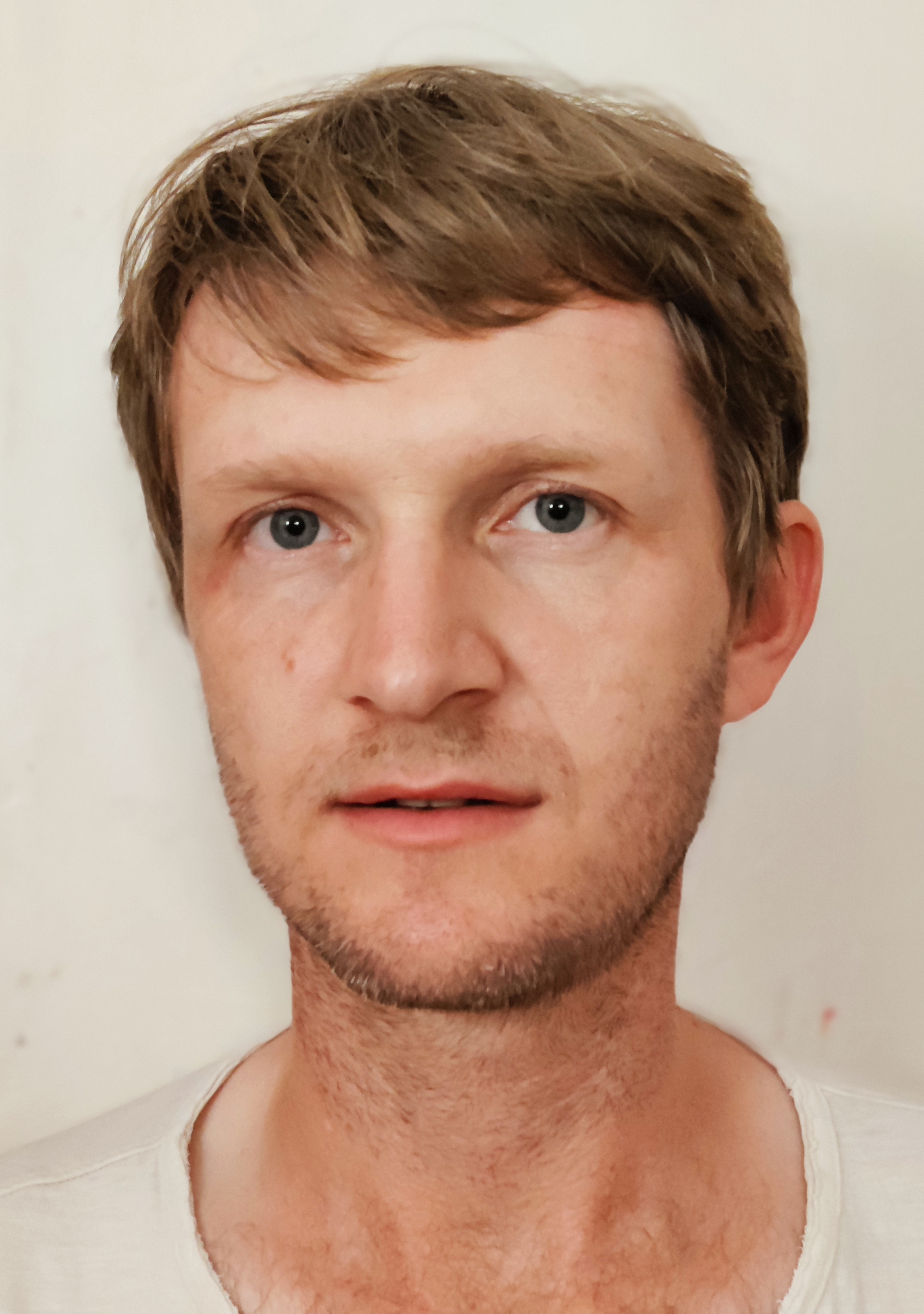}}]{Gabriel Hartmann} received a B.Sc degree in mechanical 
engineering in 2016 and an M.Sc degree in computer science in 2020. He is currently pursuing a Ph.D. degree in mechanical engineering and computer science, all at Ariel University, Ariel, Israel.

His research interest includes motion planning and control of robotics and autonomous vehicles, model-based and model-free reinforcement learning for time-optimal velocity control, and combining learning-based and classical control methods.
\end{IEEEbiography}

\begin{IEEEbiography}[{\includegraphics[width=1in,height=1.25in,clip,keepaspectratio]{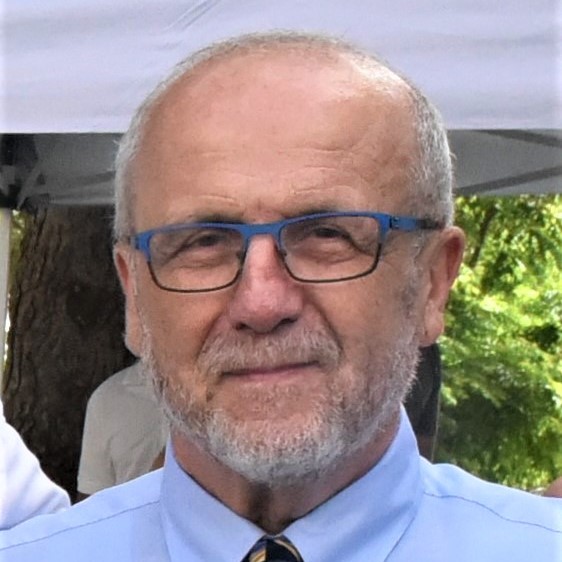}}]{Zvi Shiller}
is the founder of the Department of Mechanical Engineering and Mechatronics at Ariel University, and the director of the Paslin Laboratory for Robotics and Autonomous Vehicles.   He earned the B.Sc. engineering degree from Tel Aviv University, and the M.Sc. and Sc.D. degrees from MIT, all in Mechanical Engineering.  Professor Shiller's research activities have focused on motion planning and stability control of off-road and intelligent road vehicles, and developing a quantitative approach to resolving ethical dilemmas in autonomous driving.  With Paolo Fiorini, he developed the concept of Velocity Obstacles for motion planning in dynamic environments.   
Prof. Shiller is the founding Chair and President of the Israeli Robotics Association.   
\end{IEEEbiography}

\begin{IEEEbiography}[{\includegraphics[width=1in,height=1.25in,clip,keepaspectratio]{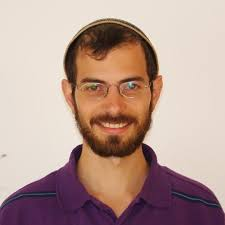}}]{Amos Azaria} is an associate professor at Ariel University, Israel. He received a B.A. degree in computer science from the Technion Institute of Technology, Haifa, Israel, in 2004, a PhD degree from Bar Ilan University, Ramat Gan, Israel in 2015 and was a post-doctoral fellow at CMU, Pittsburgh PA at the machine learning department. After completing the Bachelor's degree he spent several years in the industry, some of which included working with Microsoft R\&D Haifa, Israel. Azaria has co-authored over 70 papers, has won the Victor Lesser distinguished dissertation award for 2015, and was a member of the winning team of the DARPA SMISC competition, 2015 on bot detection. His research interests include human-agent interaction, deep learning, human aided machine learning, reinforcement learning, and natural language processing.
\end{IEEEbiography}
\EOD

\end{document}